\documentclass[10pt,twocolumn,letterpaper]{article}

\usepackage[pagenumbers]{cvpr} 

\definecolor{cvprblue}{rgb}{0.21,0.49,0.74}
\usepackage[pagebackref,breaklinks,colorlinks,allcolors=cvprblue]{hyperref}
\usepackage{graphicx}
\usepackage{bm}
\usepackage{algorithmic}
\usepackage{pgffor}
\usepackage{amsmath}
\usepackage{times}
\usepackage{multirow}
\usepackage[ruled,vlined,linesnumbered]{algorithm2e}
\usepackage[table]{xcolor}
\usepackage{booktabs}
\usepackage{array}
\usepackage{adjustbox}
\usepackage{pifont}
\usepackage[misc]{ifsym}
\usepackage[accsupp]{axessibility}  
\usepackage[ruled,vlined,linesnumbered]{algorithm2e}
\usepackage{hyperref}
\hypersetup{
    colorlinks=true,    
    linkcolor=cvprblue,     
    urlcolor=cvprblue       
}
\def\paperID{41403} 
\def\confName{CVPR}
\def\confYear{2026}
\title{POINTS-Long: Adaptive Dual-Mode Visual Reasoning in MLLMs}
\author{
Haicheng Wang$^{1,2*}$, 
Yuan Liu$^{2*}\textsuperscript{\Letter}$, 
Yikun Liu$^{1, 2*}$, 
Zhemeng Yu$^{1}$, 
Zhongyin Zhao$^{2}$, \\
Yangxiu You$^{2}$,
Zilin Yu$^{2}$,
Le Tian$^{2}$, 
Xiao Zhou$^{2}$, 
Jie Zhou$^{2}$, 
Weidi Xie$^{1}$,
Yanfeng Wang$^{1}\textsuperscript{\Letter}$ \\
$^1$ SAI, Shanghai Jiao Tong University, China \quad $^2$ WeChat AI, Tencent, China
}

\begin{document}
\maketitle
\begingroup
\renewcommand\thefootnote{\relax}\footnotetext{*: Equal contribution. \quad \textsuperscript{\Letter}: Corresponding author.}
\endgroup
\begin{abstract}
Multimodal Large Language Models (MLLMs) have recently demonstrated remarkable capabilities in cross-modal understanding and generation. However, the rapid growth of visual token sequences—especially in long-video and streaming scenarios—poses a major challenge to their scalability and real-world deployment. Thus, we introduce POINTS-Long, a native dual-mode MLLM featuring dynamic visual token scaling inspired by the human visual system. The model supports two complementary perception modes: focus mode and standby mode, enabling users to dynamically trade off efficiency and accuracy during inference. On fine-grained visual tasks, the focus mode retains the optimal performance, while on long-form general visual understanding, the standby mode retains 97.7-99.7\% of the original accuracy using only 1/40-1/10th of the visual tokens. Moreover, POINTS-Long natively supports streaming visual understanding via a dynamically detachable KV-cache design, allowing efficient maintenance of ultra-long visual memory. Our work provides new insights into the design of future MLLMs and lays the foundation for adaptive and efficient long-form visual understanding. Model and code are available at \href{https://anakin-skywalker-joseph.github.io/POINTS-Long-Webpage}{Link}.
\end{abstract}    
\vspace{-0.2cm}
\section{Introduction}
\label{sec:intro}
Multimodal Large Language Models (MLLMs)~\cite{yu2025minicpm,wang2025internvl3,bai2025qwen2,team2025kimi,coreteam2025mimovltechnicalreport,openai2024gpt4technicalreport,lu2025ovis2,yang2025kwai,v2507glm,guo2025seed1} have recently achieved remarkable progress in cross-modal comprehension and reasoning. However, these remarkable abilities come at a steep computational cost when processing long visual content like videos. The root cause lies in the visual tokenization, which expands the total sequence length with video duration, resulting in quadratic growth of computation and memory costs. This inherent scalability bottleneck remains a critical challenge for real-world long-duration applications.

\begin{figure}[t]
  \centering
  \includegraphics[width=\columnwidth]{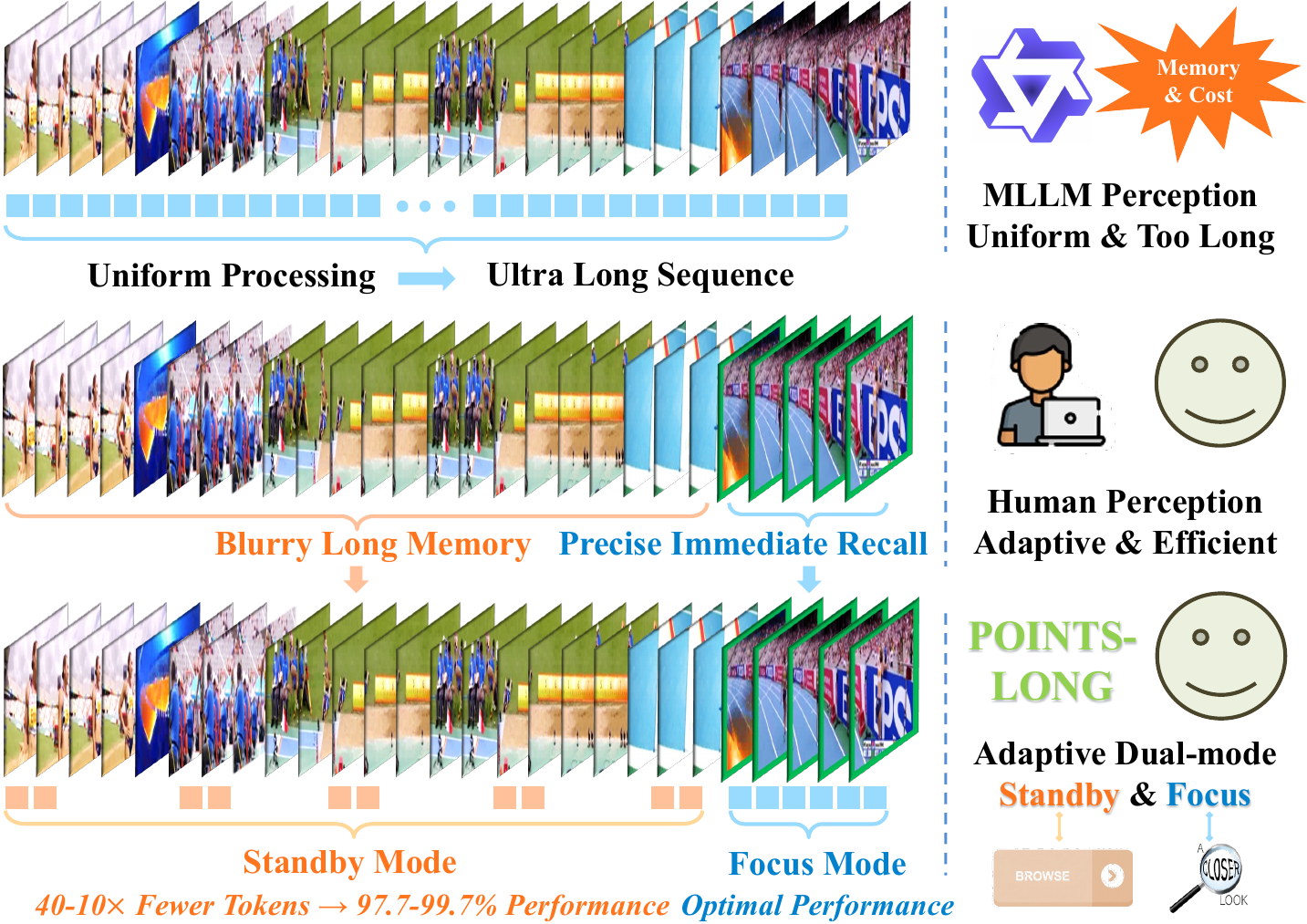}
  \vspace{-0.5cm}
  \caption{\textbf{POINTS-Long: Bridging the Gap between Human Visual Perception and MLLM Scalability.} Inspired by human's adaptive visual processing, POINTS-Long introduces a dual-mode system which switches between high-fidelity Focus Mode and efficient Standby Mode, enabling both detailed analysis and long-term streaming understanding with significantly reduced cost.}
  \label{fig:teaser}
  \vspace{-0.5cm}
\end{figure} 

Extensive research has recently yielded sophisticated strategies for visual sequence compression~\cite{zhong2024aim,liu2024multi,wen2025token}. Nevertheless, most widely-used MLLMs still rely on simple methods like pixel-shuffle~\cite{wang2024qwen2,wang2025internvl3} and pooling~\cite{guo2025seed1}. This gap between research and practice stems from three key challenges hindering the adoption of advanced techniques in production systems:
(1) Insufficient Compression Ratio: The reduction ratio is inadequate for long-video applications (thousands of frames) without a significant drop in performance~\cite{wang2025folder,tao2025dycoke,ye2025atp}. 
(2) Lack of Generality: Models are often forced into a trade-off, becoming either efficient long-video specialists~\cite{li2024videochat,shu2024video,liu2025video} that sacrifice fine-grained reasoning, or capable reasoners that cannot scale, limiting their utility as all-in-one assistants. 
(3) Deployment Difficulty: Many methods~\cite{yang2025libra,zhang2024sparsevlm,xing2024pyramiddrop} are incompatible with modern inference optimizations or frameworks (e.g., Flash-Attn~\cite{dao2023flashattention}, vLLM~\cite{kwon2023efficient}, SGLang~\cite{zheng2024sglang}), preventing their theoretical efficiency from being realized in practice. 

This suggests that a paradigm shift, rather than incremental improvements, may be necessary. We are thus motivated to ask: \textit{Is the current monotonous approach to visual processing in MLLMs inherently flawed?} We draw inspiration from the human visual system, which effortlessly processes a continuous stream of visual information without being overwhelmed. Human perception appears to operate in at least two distinct modes: a focused mode for high-fidelity details and a standby mode for low-effort, general awareness~\cite{fei2007we,vanrullen2016perceptual}. This duality is also reflected in our hierarchical memory~\cite{baddeley2003working,luck1997capacity}: precise immediate recall, blurry short-term memories, and semantic long-term recollections, like a textual summary. This reveals an efficient architecture: a precise buffer for the present, a compressed cache for short-term, and a conceptual archive for long-term.

Inspired by this human cognitive model, we introduce POINTS-Long, a MLLM built upon POINTS1.5~\cite{liu2024points1}. Its core innovation is a native dual-mode visual processing system:
\textbf{Focus Mode}: Uses the complete visual sequence for tasks requiring fine-grained analysis, ensuring maximum performance.
\textbf{Standby Mode}: Operates on a drastically reduced number of visual tokens for the holistic perception of long videos, with only a negligible drop in performance.

To implement this functionality without compromising the model's original strengths, we employ a two-stage post-training adaptation process. First, in a visual distillation stage, we freeze the original MLLM and train a small set of new parameters to distill the rich information from the full visual sequence into a compact set of "Standby tokens" (Sec.~\ref{sec:arch}). This ensures that the Standby tokens are semantically aligned with the original "Focus tokens" while leaving the Focus mode's pathway entirely unaffected. In the second stage, we adapt the LLM by fine-tuning it with a small learning rate on high-quality data, enabling it to effectively interpret inputs from both modes (Sec.~\ref{sec:stage}).

This strategic approach yields remarkable efficiency: on the OpenCompass video benchmark, our Standby mode retains 97.7\%–99.7\% of the original model's performance while using just 1/40th to 1/10th of the visual tokens. Crucially, this efficiency is achieved without compromise, as the Focus mode fully preserves the model's original fine-grained capacity. Furthermore, this dual-mode architecture enables a more effective approach to streaming vision. By dynamically combining modes, POINTS-Long emulates a human-like memory system—a high-fidelity "present" (Focus) and a compressed "short-term" (Standby)—through a novel detachable KV cache mechanism. This allows for native, long-term understanding without costly context re-prefills. Notably, POINTS-Long is designed for practical deployment; all evaluations were conducted using SGLang~\cite{zheng2024sglang} inference framework. Overall, our contributions can be summarized as follows:
\begin{itemize}
\item We introduce POINTS-Long, a novel MLLM inspired by human cognition. It features a dual-mode visual system (Focus and Standby) that resolves the critical trade-off between fine-grained reasoning and long-vision scalability.

\item We propose a generalizable two-stage post-training strategy that can efficiently equip a well-trained MLLM with the high-compression Standby mode while fully preserving its original performance in the Focus mode.

\item We demonstrate the practical viability and state-of-the-art efficiency of our approach. POINTS-Long natively supports long-term streaming video understanding through a novel detachable KV cache mechanism and is fully compatible with modern inference frameworks, achieving up to 6.2$\times$ generation throughput with negligible loss.
\end{itemize}
\section{Related Work}
\label{sec:related}

\noindent
\textbf{Video Large Language Models.} 
MLLMs have demonstrated impressive capabilities in understanding multimodal information like video~\cite{li2024llava,wang2025internvl3,bai2025qwen2,team2025kimi,coreteam2025mimovltechnicalreport,zhang2025videollama,li2023videochat,yang2025kwai,v2507glm,guo2025seed1}. However, the rapid growth in computational cost from the large number of visual tokens severely limits their scalability for practical, long-form video tasks. 
To address this bottleneck, some MLLMs~\cite{li2024videochat,shu2024video,liu2025video,tao2025dycoke} employ visual token compression for efficient long-form understanding. However, they often result in highly specialized models: some sacrifice fine-grained image reasoning to become video experts, while others~\cite{qian2024streaming,zhang2025flash} built for streaming video are even more task-specific. This specialization highlights a critical need for a native MLLM that can perform both long-video processing and precise image analysis.

\noindent
\textbf{Efficient MLLMs Inference.} 
The practical deployment of MLLMs is dominated by inference frameworks like vLLM~\cite{kwon2023efficient} and SGLang~\cite{zheng2024sglang}. These systems achieve state-of-the-art throughput by leveraging kernel-level optimizations like FlashAttention~\cite{dao2023flashattention} and PagedAttention~\cite{kwon2023efficient}. However, many visual token reduction methods are incompatible with these frameworks (or hard to implement), {\em e.g., } requiring explicit attention matrices or disrupting the uniform block structure of the KV cache. As a result, their theoretical efficiency doesn't translate to real-world performance, severely limiting their practical use. 

\noindent
\textbf{Visual Token Reduction in MLLMs.} Some preliminary studies mainly focus on Vision Transformers~\cite{rao2021dynamicvit,kong2022spvitenablingfastervision,bolya2022token} and KV cache compression~\cite{zhang2023h2o,li2024snapkv,sun2024shadowkv} for LLMs. In the context of MLLMs, common methods like Q-Former~\cite{li2023blip}, resampler~\cite{chen2024internvl} and pooling~\cite{chen2023minigpt} are widely used during the training phase to reduce visual tokens. Recently, some studies tried to handle the token reduction problem in more delicate ways~\cite{shang2024llava,alvar2025divprune,huang2024prunevid,huang2024dynamic,wen2025efficient,wan2024look,huang2024accelerating,ye2025voco}. In particular, training-free methods mainly leverage task-orientated attention importance~\cite{chen2024image,zhang2024sparsevlm,xing2024pyramiddrop,liu2025compression}, or inherent visual redundancy~\cite{ju2024turbo,wang2025folder,yang2025visionzip}, compromising efficiency with performance. Methods that require additional training~\cite{li2024mini,li2025tokenpacker,shi2024we,li2024llama,zhang2025llava} can compress visual tokens more effectively, but they often enforce a fixed trade-off, leading to performance degradation and poor extensibility. We aim to build a natively adaptive MLLM that provides the flexibility to dynamically balance between computational efficiency and reasoning accuracy.

\section{Method}
\subsection{Overview}
\label{sec:overview}
Our dynamic visual understanding framework is inspired by the human visual system, incorporating both a Focus Mode and a Standby Mode. This design aims to selectively and drastically reduce computational load, and potentially maintain long visual memory, which is guided by four key principles: 
(P1) Performance Preservation: The Focus Mode remains equivalent to the original, well-trained MLLM.
(P2) Optimized Standby Performance: The Standby Mode strives to approximate Focus Mode quality with drastically lower cost.
(P3) Deployment Simplicity: The architecture should be easy to deploy and compatible with modern inference frameworks (e.g., vLLM~\cite{kwon2023efficient}, SGLang~\cite{zheng2024sglang}) for real-world speed-ups.
(P4) Extensibility: The training solution should be adaptable to a wide variety of existing MLLMs.

To adhere to these principles, we begin with an instruct model and introduce the Standby Mode capacity via one post-training phase. We add several learnable modules between the vision backbone and the projector (Sec.~\ref{sec:compress}), a solution designed to satisfy (P1) and (P3) while maximizing the performance of (P2). 
We then propose a two-stage training strategy, including (1) Visual Distillation and Alignment (2) LLM Mode Adaptation, to efficiently integrate this new mode (Sec.~\ref{sec:stage}). The resulting model natively supports both focus and standby modes. This dual-mode capability, with a novel detachable KV cache mechanism, allows the model to naturally support efficient, streaming visual understanding (Sec.~\ref{sec:inference}) while keeping its full capacity.

\subsection{Architecture of Base MLLM}
\label{sec:arch}
We use POINTS1.5-8B-Instruct (improved version of POINTS1.5~\cite{liu2024points1}) for experiments, a highly competitive MLLM comparable to mainstream MLLMs like Qwen2.5-VL~\cite{bai2025qwen2}. It is composed of a LLM initialized from Qwen3-8B-base~\cite{yang2025qwen3} (1D RoPE~\cite{su2024roformer} for visual inputs) and a native-resolution image encoder initialized from Qwen2-VL-ViT~\cite{wang2024qwen2} (employing 2D RoPE). This base model has already undergone a comprehensive, multi-stage training pipeline, including multimodal alignment, continued pretraining, multimodal SFT, and post-training phase. (details are in supplementary material). Our proposed dynamic dual-mode scheme is applied as a post-training phase on top of this instruct model. Note that our approach can be applied to any MLLM following a similar architecture.

\begin{figure*}
  \centering
  \includegraphics[width=0.98\textwidth]{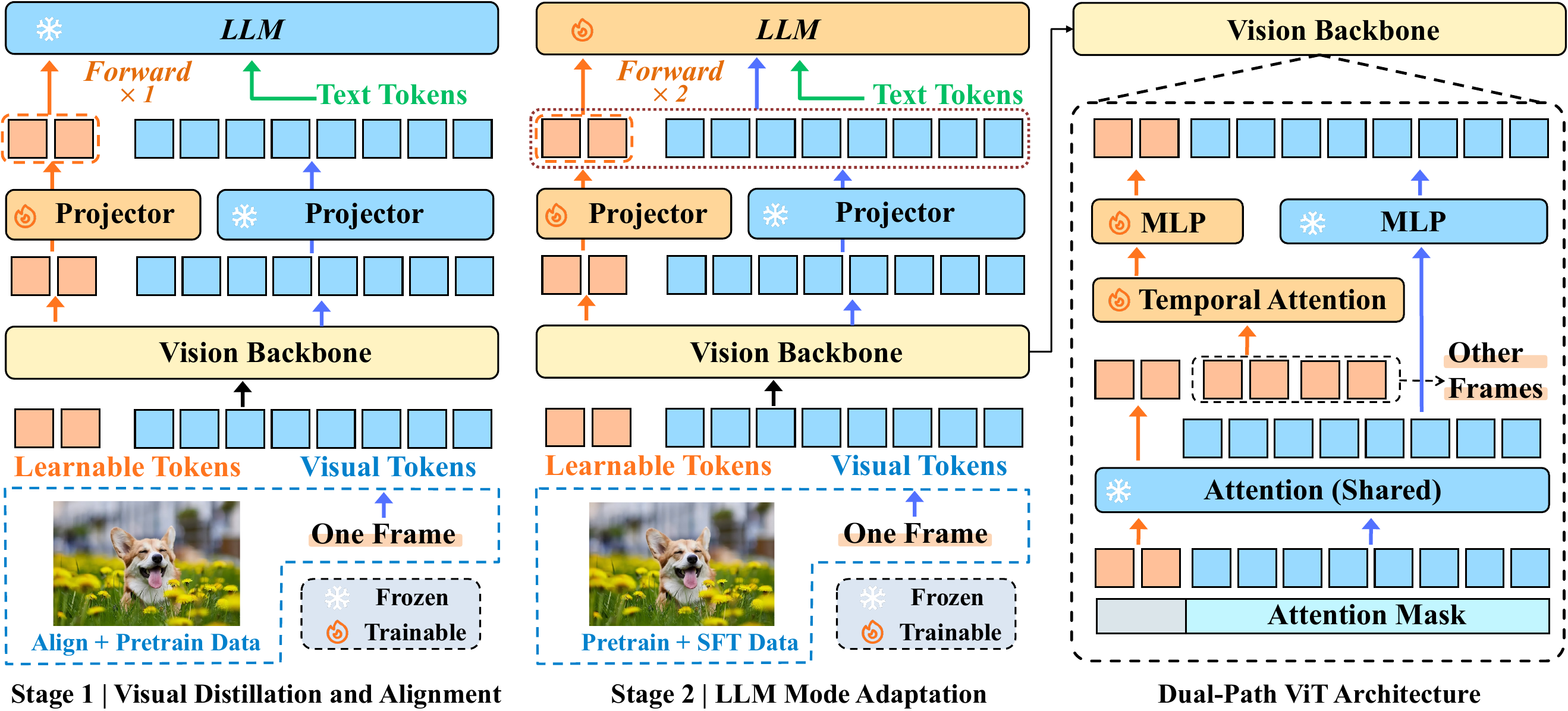}
  \vspace{-0.1cm}
  \caption{\textbf{POINTS-Long Architecture.} The original visual patch sequence (blue) is processed by the original ViT modules. We introduce $n$ learnable tokens (orange) processed through duplicated learnable MLPs and projector, to act as the compressed representation of the full sequence. An additional temporal modeling allows better compression for video inputs. With symmetric attention mask, the original path is totally unaffected, thus preserving its performance. This dual-mode system is enabled by a two-stage post-training: Stage 1 (left) trains only the new parameters for visual distillation, while Stage 2 (middle) fine-tunes the LLM with a small learning rate for mode adaptation.}
  \label{fig:pipeline}
  \vspace{-0.3cm}
\end{figure*}

\begin{figure}
  \centering
  \includegraphics[width=0.99\columnwidth]{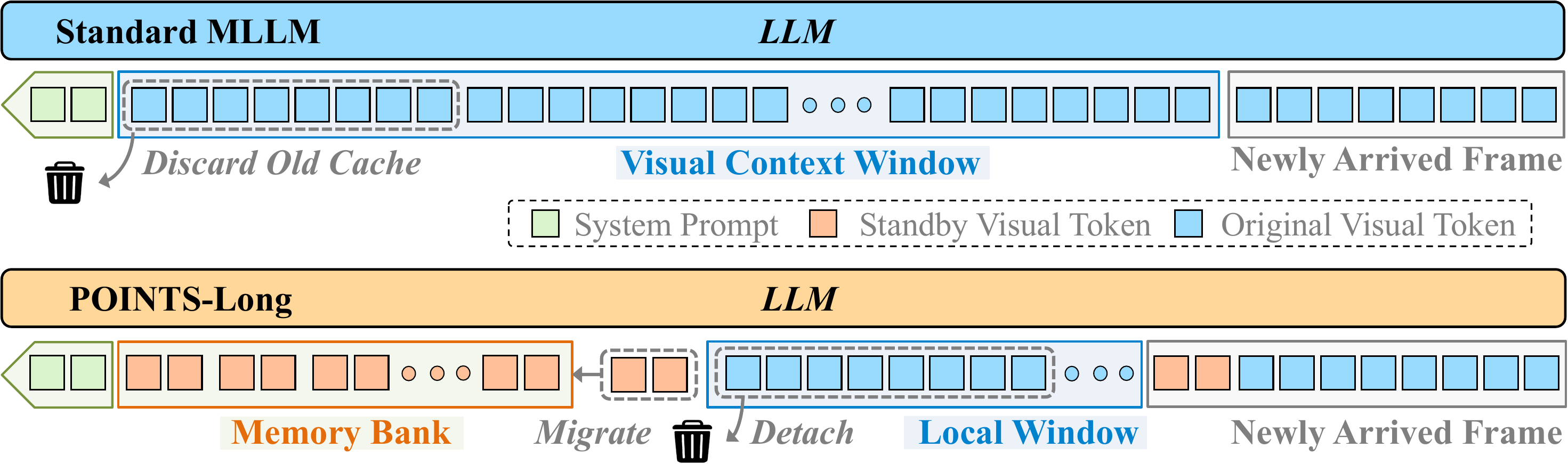}
  \vspace{-0.2cm}
  \caption{\textbf{Streaming Inference in LLM.} ($\uparrow$) When handling streaming inputs, general MLLMs discard previous cached context when reaching maximum budget. ($\downarrow$) POINTS-Long encodes new frames in Focus Mode. When local window is full, the original sequence's cache is detached, and the compact standby-sequence cache is migrated to a long-term "Memory Bank".}
  \label{fig:stream}
  \vspace{-0.5cm}
\end{figure}

\subsection{Native Visual Compression Structure}
\label{sec:compress}
Starting from POINTS1.5-8B-Instruct architecture, we introduce a novel modification to the vision backbone (ViT) and projector, keeping the original inference path unchanged. The core objective is to distill the vast information from the original visual sequence into a small set of tokens, enabling a highly efficient "standby" mode without compromising the performance of the original "focus" mode.
\subsubsection{Dual-Path ViT Architecture}
Inspired by CLIP~\cite{radford2021learning}, we append $n$ learnable tokens onto the patchified sequence, where $n$ is significantly smaller than the average visual sequence length. These tokens are intended to act as a compressed representation of the full sequence. However, integrating these learnable tokens introduces a training dilemma: (1) If we freeze the ViT and train only the learnable tokens, the model lacks the fitting capability to distill complex visual information, leading to poor performance (Tab.~\ref{tab:ablation}). (2) Unfreeze ViT can improve the fitting capability, but the training dynamics are altered, impairing the model's original "focus mode" performance.

To resolve this, we re-architect the ViT by introducing a parallel processing path for new learnable tokens, shown in Fig~\ref{fig:pipeline}. Similar to MoT~\cite{deng2025emerging}, for each MLP layer in the ViT, we duplicate it to create a new one, which is initialized with the weights of the original MLP. The original visual sequence is processed by the original MLPs, while the new learnable tokens are processed exclusively by these new MLPs. This parallel structure is also mirrored in the final projector with the same operation. The key interaction between these two paths is the shared attention block. This simple design significantly boosts the performance (Tab.~\ref{tab:ablation}).

In addition, to preserve the invariability of the original "focus" path, we employ an asymmetric attention mask: the original patch tokens compute attention only among themselves (masking out new learnable tokens), ensuring the invariance of their representations. In contrast, the learnable tokens are allowed to attend to the entire sequence, enabling them to aggregate global visual information. This simple masking strategy is fully compatible with Flash Attention~\cite{dao2023flashattention}. Finally, we assign positional embeddings to the learnable tokens by uniformly sampling the original 2D RoPE~\cite{su2024roformer}, an initialization technique that, as visualized in our supplementary material, encourages different tokens to specialize in different spatial regions of the image.

\noindent \textbf{Discussion}
This methodology was designed to achieve the 4 principles outlined in Sec.~\ref{sec:overview}: The parallel ViT architecture and asymmetric attention mask ensure the original path is undisturbed, maintaining focus mode performance and easing deployment. The added parameters enhance the representation ability, boosting the capacity of standby mode.

\subsubsection{Temporal Modeling}
The architecture so far originates from POINTS1.5 image encoder and, consequently, solely addresses intra-frame spatial redundancy, overlooking the significant temporal redundancy in video inputs, which can be more critical.

A naive application of our method would compress each frame into $n$ tokens independently and then concatenate. However, a joint compression that models spatio-temporal relationships could achieve higher information fidelity. Therefore, to further enhance the standby mode's efficacy for video, we introduce an explicit temporal modeling component. As shown in Fig~\ref{fig:pipeline}, we insert a temporal attention module into the final layers (last 5) of the ViT, positioning it between attention and MLP blocks. This module operates only on the compressed learnable token sequences. It concatenates the learnable tokens from $k$ adjacent frames and applies causal attention across this new temporal sequence. We use standard 1D RoPE as position encoding.

Through this temporal attention layer, the compressed representations of neighboring frames can exchange and refine information, significantly raising the upper bound of information retention in the final standby sequence for video understanding. The use of causal attention is a deliberate design choice to ensure compatibility with streaming video encoding scenarios (as detailed in the supplementary material). We note that this explicit temporal module is only designed for MLLMs with image encoder as ViT, while for those using native video encoder, it's no longer necessary.

\subsubsection{Two-Stage Dual-Mode Training}
\label{sec:stage}
To adapt the MLLM to these two distinct modes, we propose a two-stage training pipeline (Fig.~\ref{fig:pipeline}).

\noindent \textbf{Stage 1: Visual Distillation and Alignment}
We freeze all parameters of the original POINTS1.5 model (ViT, projector, and LLM) and train only the newly introduced components: the learnable tokens, the duplicated MLP, projector and the temporal attention layers. During this stage, the LLM is fed only the compressed learnable token sequence. This stage functions similarly to the alignment phase in MLLM training, forcing the new modules to distill the essential visual information into the compact token sequence. For this stage, we use the POINTS1.5 alignment data and a subset of the multimodal continue-pretrain data.

\noindent \textbf{Stage 2: LLM Mode Adaptation}
After Stage 1, the learnable tokens effectively carry the distilled visual information. However, the LLM has not been trained to understand this new, compressed sequence format. Therefore, in Stage 2, we unfreeze the LLM and fine-tune it with a small learning rate, training it jointly with the Stage 1 parameters. Meanwhile, a critical challenge arises: even low-LR fine-tuning can degrade the LLM's performance on the original focus mode. To mitigate this, we employ a 2-forward training strategy: In each training step, we perform two forward passes. \textit{Pass 1 (Standby)}: We feed LLM the short learnable token sequence. \textit{Pass 2 (Focus)}: We feed the LLM the full sequence (learnable tokens + original tokens).

We average the losses from both passes and backpropagate the combined loss. This joint objective forces the LLM to adapt to the new standby mode while maintaining its original focus mode capabilities. As shown in Tab.~\ref{tab:image} and Tab.~\ref{tab:ablation}, this method significantly improves standby mode performance while fully preserving focus mode accuracy.

\noindent \textbf{Discussion}
While other token compression modules exist, e.g. resampler~\cite{jaegle2021perceiver} or Q-Former~\cite{li2023blip}, they often suffer from training instability due to the random initialization of new parameters. Our approach, by contrast, initializes all new modules from the pre-trained weights, ensuring a more stable training. Furthermore, our parallel MLP design maintains better computational parallelism than sequential cross-attention modules. Still, our primary contribution is not the specific compression module itself, but the introduction of a human-like, dual-mode paradigm that allows a model to switch between high-efficiency (standby) and high-fidelity (focus) visual processing at will.

\subsection{Model Inference}
\label{sec:inference}
\noindent \textbf{Offline Inference} Following the two-stage training, POINTS-Long can perform two distinct modes, Focus and Standby, which can be selected based on task requirements.

\noindent (1) Fine-grained Understanding (Focus Mode): For tasks demanding high-fidelity detail, Focus Mode is employed to achieve optimal performance (see Tab.~\ref{tab:image}). 

\noindent (2) Holistic Long-sequence Understanding (Standby Mode): For tasks involving holistic comprehension or long visual sequences (e.g., video-QA), we switch to Standby Mode. This mode achieves nearly identical performance using drastically fewer tokens. For example, processing 64 frames of a 480p video, which originally required $\approx$20k visual tokens, now requires only 0.5k-2k tokens while retaining 97.7-99.7\% of the full-sequence performance. Notably, Standby Mode effectively overcomes the context length limitations of most MLLMs (e.g., 32k). By representing each frame compactly, POINTS-Long can gain steadily with respect to sampled frame number (Tab.~\ref{tab:frame}).

\noindent \textbf{Streaming Inference} POINTS-Long is inherently well-suited for the streaming scenario. Previous models face a critical limitation on streaming understanding: as new frames are encoded (prefilled), the context limit/KV cache budget will eventually be reached. At that time the oldest cached frame will be discarded, resulting in a short memory window, e.g., prefilling a 480p video at 2fps would retain only about 50 seconds of visual memory for 32K context.

Meanwhile, POINTS-Long enables a far more effective hybrid memory strategy. As shown in Fig~\ref{fig:stream}, we can maintain a "local window" by Focus Mode (prefilling new frames using short+full sequence) and a "memory bank" in Standby Mode (retaining only the short-seq KV cache from older frames). When the local window limit is reached, we only discard the large full-sequence cache, migrating its compact standby-sequence cache into the long-term memory bank. This allows us to manage a 32k context budget dynamically, e.g., a 4k local window and a 28k memory bank would allow the model to maintain ~6 seconds of complete current visual information (Focus) while preserving up to 30 minutes of compressed visual memory (Standby). This represents up to 40x increase in memory duration.

\noindent \textbf{Discussion}
In real-world scenarios, many tasks prioritize efficiency over fine-grained detail, e.g., video tagging and security auditing. Concurrently, emerging applications like interactive livestreaming and multimodal assistants demand both long-term comprehension and high-fidelity analysis. Meanwhile in MLLM design, efficiency and granularity are always treated as a fixed trade-off, {\em i.e.}, either an efficient model or a performant one. Inspired by the observations of human visual processing, we argue that the two modes should be decoupled and fit in one single model, {\em i.e.}, rather than a \textbf{fixed trade-off}, they should represent a \textbf{choice}.

In this work, the choice between modes is predefined at inference time. We believe that an advanced model could learn to make this choice dynamically——learning which parts of a video to "glance" at (Standby) and which to "scrutinize" (Focus). This concept, which we term "Thinking with Videos"~\cite{zhang2025thinking}, will be explored in future work.
\section{Experiments}

\begin{table*}[t]
\centering
\footnotesize
\setlength{\tabcolsep}{5pt}
\caption{\textbf{Opencompass Video Benchmark.} Under the same setting (64 frames), POINTS-Long achieves competitive performance (97.7\%-99.7\%) against original POINTS1.5-8B with drastically less tokens (2.5\%-10\%), and retains even better focus mode performance.}
\vspace{-0.3cm}
\resizebox{0.99\textwidth}{!}{
\begin{tabular}{l|c|c|c|ccccccl}
\toprule
\textbf{Model} & \begin{tabular}[c]{@{}c@{}} \textbf{Num} \\ \textbf{Frame} \end{tabular} &\begin{tabular}[c]{@{}c@{}} \textbf{Token/} \\ \textbf{Frame}\end{tabular} &\begin{tabular}[c]{@{}c@{}}\textbf{Total Num}\\ \textbf{of Token}\end{tabular} & \textbf{MVBench} & \textbf{Video-MME} & \textbf{MMBench-Video} & \textbf{Tempcompass} & \textbf{MLVU} & \textbf{LongVideoBench} & \textbf{Avg} \\ \hline  \hline
Qwen2-VL-7B~\cite{wang2024qwen2} & 64 & - & $>$12k & 66.0 & 59.7 & 48.3 & 69.6 & 66.4 & 55.6 & 60.9 \\
MiMo-VL-7B-RL~\cite{coreteam2025mimovltechnicalreport} & 64 & - & $>$12k & 63.2 & 65.0 & 54.0 & - & 66.2 & - & - \\
Qwen2.5-VL-7B~\cite{bai2025qwen2} & 64 & - & $>$12k & 67.5 & 62.8 & 53.0 & 72.0 & 70.2 & 56.0 & 63.6 \\
VideoLLaMA3-7B~\cite{zhang2025videollama} & - & - & $>$12k & 69.7 & 66.2 & - & 68.1 & 73.0 & 59.8 & - \\
InternVL2.5-8B~\cite{chen2024expanding} & 64 & - & $>$12k & 70.5 & 63.7 & 56.0 & 68.7 & 68.5 & - & - \\
Qwen2.5-Omni-7B~\cite{xu2025qwen2} & 64 & - & $>$12k & 69.0 & 64.1 & 55.0 & 70.7 & 67.5 & - & - \\ 
InternVL3-8B~\cite{zhu2025internvl3} & 64 & - & $>$12k & 73.2 & 66.0 & 56.3 & 70.4 & 71.4 & 58.8 & 66.0 \\ 
GLM4.1V-9B~\cite{v2507glm} & - & - & $>$12k & 68.2 & 68.4 & 54.3 & - & 71.5 & 65.7 & - \\
Kimi-VL-A3B-2506~\cite{team2025kimi} & - & - & $>$12k & 59.7 & 67.8 & - & - & 74.2 & 64.5 & - \\
InternVL3.5-8B~\cite{wang2025internvl3} & - & - & $>$ 12k & 72.1 & 66.0 & 55.7 & - & 70.2 & 62.1 & - \\ \hline
\rowcolor{cyan!10} POINTS1.5-8B (baseline) & 64 & 324 & $\approx$ 20K & 60.3 & 66.1 & 61.0 & 71.1 & 72.0 & 59.8 & 65.0 \\
POINTS1.5-8B (low-res) & 64 & 32 & 2048 & 54.9 & 61.2 & 51.0 & 67.1 & 67.3 & 53.9 & 59.2 \\
POINTS1.5-8B (pooling) & 64 & 32 & 2048 & 54.9 & 55.4 & 43.0 & 66.6 & 67.1 & 54.5 & 56.9 \\
\rowcolor{orange!10} POINTS-Long (standby) & 64 & 8 & 512 (2.5\%) & 59.4 & 63.5 & 58.0 & 69.9 & 71.9 & 58.2 & 63.5 (97.7\%) \\
\rowcolor{orange!10} POINTS-Long (standby) & 64 & 16 & 1024 (5\%) & 59.7 & 65.0 & 59.3 & 69.1 & 71.7 & 58.9 & 63.9 (98.3\%) \\
\rowcolor{orange!10} POINTS-Long (standby) & 64 & 32 & 2048 (10\%) & 60.8 & 65.7 & 60.9 & 70.3 & 71.6 & 59.5 & 64.8 (99.7\%) \\
\rowcolor{orange!10} POINTS-Long (focus) & 64 & 324+32 & $\approx$ 22K & 61.0 & 66.1 & 60.3 & 71.3 & 73.2 & 59.4 & 65.2 (100.3\%) \\
\bottomrule
\end{tabular}
}
\vspace{-0.1cm}
\label{tab:opencompass-video}
\end{table*}

\begin{table*}[t]
\centering
\footnotesize
\setlength{\tabcolsep}{3pt}
\caption{\textbf{More Video Benchmarks.} We evaluate on more video benchmarks to prove universality.}
\vspace{-0.3cm}
\resizebox{0.99\textwidth}{!}{
\begin{tabular}{l|c|c|c|cccccccl}
\toprule
\textbf{Model} & \begin{tabular}[c]{@{}c@{}}\textbf{Num} \\ \textbf{Frame} \end{tabular} &\begin{tabular}[c]{@{}c@{}}\textbf{Token/} \\ \textbf{Frame}\end{tabular} &\begin{tabular}[c]{@{}c@{}}\textbf{Total Num}\\ \textbf{of Token}\end{tabular} & \begin{tabular}[c]{@{}c@{}}\textbf{CG-Bench} \\ \textbf{(long-acc)} \end{tabular} & \textbf{MovieChat1k} & \textbf{Egoschema} & \textbf{LVBench} & \textbf{TemporalBench} & \textbf{Activitynet-qa} & \textbf{WorldSense} & \textbf{Avg} \\ \hline  \hline
Moviechat~\cite{song2024moviechat} & - & - & $>$12k & - & 62.3 & 53.5 & 22.5 & - & 45.7 & - & - \\
LLaVA-OV-7B~\cite{li2024llava} & 64 & - & $>$12k & 30.9 & - &  59.8 & 26.9 & 59.4 & 56.0 & 37.7 & - \\
MiMo-VL-7B-RL~\cite{coreteam2025mimovltechnicalreport} & 64 & - & $>$12k & - & - & 59.4 & 37.1 & - & - & - \\
LLaVA-Video-7B~\cite{zhang2024video} & - & - & $>$12k & - & - & 57.3 & - & 63.6 & 56.5 & 40.2 & - \\
InternVL3-8B~\cite{zhu2025internvl3} & - & - & $>$12k & 38.6 & - & - & 44.1 & - & - & - & - \\ \hline
\rowcolor{cyan!10} POINTS1.5-8B (baseline) & 64 & 324 & $\approx$ 20K & 36.7 & 77.0 & 60.6 & 44.3 & 64.3 & 54.3 & 40.4 & 53.9 \\
\rowcolor{orange!10} POINTS-Long (standby) & 64 & 8 & 512 (2.5\%) & 33.6 & 76.0 & 59.7 & 40.4 & 64.4 & 55.4 & 39.4 & 52.7 (97.8\%) \\
\rowcolor{orange!10} POINTS-Long (standby) & 64 & 16 & 1024 (5\%) & 34.6 & 73.0 & 60.5 & 42.5 & 65.1 & 55.1 & 39.7 & 52.9 (98.1\%)  \\
\rowcolor{orange!10} POINTS-Long (standby) & 64 & 32 & 2048 (10\%) & 35.7 & 77.0 & 62.6 & 42.6 & 64.1 & 54.7 & 40.0 & 53.8 (99.8\%) \\
\rowcolor{orange!10} POINTS-Long (focus) & 64 & 324+32 & $\approx$ 22K & 35.4 & 79.0 & 62.5 & 44.5 & 64.5 & 54.7 & 40.4 & 54.4 (100.9\%) \\
\bottomrule
\end{tabular}
}
\label{tab:video-bench}
\vspace{-0.3cm}
\end{table*}

\begin{table*}[t]
\centering
\footnotesize
\caption{\textbf{Scalability of Inference Frame.} By drastically compressing visual tokens, POINTS-Long can process more frames without context length overflow. we witness a steady gain with respect to frame number, which is not always the case for general MLLMs.}
\vspace{-0.2cm}
\resizebox{0.99\textwidth}{!}{
\begin{tabular}{l|c|c|c|cccccccc}
\toprule
\textbf{Model} & \begin{tabular}[c]{@{}c@{}}\textbf{Num} \\ \textbf{Frame} \end{tabular} &\begin{tabular}[c]{@{}c@{}}\textbf{Token/} \\ \textbf{Frame}\end{tabular} &\begin{tabular}[c]{@{}c@{}}\textbf{Total Num}\\ \textbf{of Token}\end{tabular} & \textbf{LVBench} & \begin{tabular}[c]{@{}c@{}}\textbf{VideoMME}\\ \textbf{(Long/Overall)}\end{tabular} & \textbf{MMBench-Video} & \begin{tabular}[c]{@{}c@{}}\textbf{CG-Bench}\\ \textbf{(60+/Overall)}\end{tabular} & \textbf{MLVU} & \begin{tabular}[c]{@{}c@{}}\textbf{LongVideoBench}\\ \textbf{(3600+/Overall)}\end{tabular} & \textbf{Avg} \\ \hline  \hline
\rowcolor{cyan!10} POINTS1.5-8B & 64 & 324 & $\approx$ 20K & 44.3 & 56.0/66.1 & 61.0 & 31.1/36.7 & 72.0 & 50.7/59.8 & 52.5 \\
\rowcolor{cyan!10} POINTS1.5-8B & 128 & 144 & $\approx$ 18K & 45.4 & 54.4/65.0 & 61.3 & 32.4/37.0 & 72.0 & 51.2/60.2 & 52.8 \\ \hline
\rowcolor{orange!10}POINTS-Long & 64 & 8 & 512 (2.5\%) & 40.4 & 54.0/63.5 & 58.0 & 29.5/33.6 & 71.9 & 49.3/58.2 & 50.5 \\
\rowcolor{orange!10}POINTS-Long & 128 & 8 & 1024 (5\%) & 42.9 & 56.4/64.4 & 59.0 & 31.1/35.3 & 71.9 & 49.5/59.6 & 51.8 \\
\rowcolor{orange!10}POINTS-Long & 256 & 8 & 2048 (10\%) & 43.6 & 57.1/66.1 & 60.0 & 30.4/36.0 & 72.4 & 50.7/59.7 & 52.4 \\ \hline
\rowcolor{orange!10}POINTS-Long & 64 & 16 & 1024 (5\%) & 42.5 & 55.3/65.0 & 59.3 & 30.7/34.6 & 71.7 & 48.4/58.9 & 51.3 \\
\rowcolor{orange!10}POINTS-Long & 128 & 16 & 2048 (10\%) & 43.1 & 56.4/66.4 & 61.0 & 33.2/36.2 & 72.7 & 51.6/60.3 & 53.0 \\
\rowcolor{orange!10}POINTS-Long & 256 & 16 & 4096 (20\%) & 44.1 & 58.0/66.9 & 61.3 & 33.6/37.4 & 72.2 & 50.7/59.5 & 53.3 \\ \hline
\rowcolor{orange!10}POINTS-Long & 64 & 32 & 2048 (10\%) & 42.6 & 55.9/65.7 & 60.9 & 32.0/35.7 & 71.6 & 48.6/59.5 & 51.9 \\
\rowcolor{orange!10}POINTS-Long & 128 & 32 & 4096 (20\%) & 45.3 & 56.9/66.9 & 62.0 & 32.0/37.3 & 72.5 & 51.1/60.4 & 53.3  \\
\rowcolor{orange!10}POINTS-Long & 256 & 32 & 8192 (40\%) & 46.9 & 58.0/66.5 & 61.3 & 34.4/37.4 & 72.5 & 49.8/59.8 & 53.8 \\
\bottomrule
\end{tabular}
}
\label{tab:frame}
\end{table*}

\begin{table*}[t]
\centering
\footnotesize
\setlength{\tabcolsep}{7pt}
\caption{\textbf{Opencompass Image Benchmark.} We show that our two-stage training will not harm the fine-grained capacity of focus mode. Bonus: With simple training-free attention-based pruning, the focus mode can be more efficient, beating other training-free baselines.}
\vspace{-0.2cm}
\resizebox{0.99\textwidth}{!}{
\begin{tabular}{l|cccccccccc}
\toprule
\textbf{Model} & \textbf{MMBench} & \textbf{MMStar} & \textbf{MMMU\_val} & \textbf{MathVista} & \textbf{OCRBench} & \textbf{AI2D} & \textbf{HallusionBench} & \textbf{MMVet} & \textbf{Avg} \\ \hline  \hline
Qwen2-VL-7B~\cite{wang2024qwen2} & 81.0 & 60.7 & 53.7 & 61.6 & 84.3 & 83.0 & 50.4 & 61.8 & 67.1 \\
InternVL2.5-8B~\cite{chen2024expanding} & 82.5 & 63.2 & 56.2 & 64.5 & 82.1 & 84.6 & 49 & 62.8 & 68.1 \\
MiniCPM-o-2.6~\cite{yao2024minicpm} & 80.6 & 63.3 & 50.9 & 73.3 & 88.9 & 86.1 & 51.1 & 67.2 & 70.2 \\
Qwen2.5-VL-7B~\cite{bai2025qwen2} & 82.2 & 64.1 & 58 & 68.1 & 88.8 & 84.3 & 51.9 & 69.7 & 70.9 \\
Ovis2-8B~\cite{lu2025ovis2} & 83.6 & 64.6 & 57.4 & 71.8 & 89.1 &86.6 & 56.3 & 65.1 & 71.8 \\ 
SAIL-VL1.6-8B~\cite{yin2025sail} & 84.0 & 69.5 & 55.4 & 74.2 & 90.5 & 87.5 & 54.4 & 73.3 & 73.6 \\ 
InternVL3-8B~\cite{zhu2025internvl3} & 82.1 & 68.7 & 62.2 & 70.5 & 88.4 & 85.1 & 49.0 & 82.8 & 73.6 \\ \hline
\rowcolor{cyan!10} POINTS1.5-8B (baseline) & 81.9 & 65.7 & 53.2 & 70.9 & 85.8 & 83.9 & 50.1 & 64.7 & 69.5 \\
\rowcolor{orange!10} POINTS-Long (focus) & 82.1 & 66.1 & 53.7 & 70.6 & 85.5 & 84.2 & 48.3 & 66.7 & 69.7 \\ \hline
\rowcolor{orange!10} + Attn-prune 50\% & 82.0 & 64.5 & 52.4 & 69.0 & 83.5 & 84.2 & 47.2 & 66.7 & 68.7 \\
+ Avg-pooling 50\% & 80.7 & 62.2 & 53.0 & 64.6 & 75.1 & 83.8 & 48.5 & 65.5 & 66.7 \\
+ Folder~\cite{wang2025folder} 50\% & 81.5 & 64.0 & 52.7 & 64.0 & 75.5 & 83.5 & 49.1 & 64.9 & 66.9 \\
\bottomrule
\end{tabular}
}
\label{tab:image}
\vspace{-0.3cm}
\end{table*}

\begin{table*}[t]
\centering
\footnotesize
\setlength{\tabcolsep}{5pt}
\caption{\textbf{Streaming Understanding.} General MLLMs struggle at long-range streaming VQA, while POINTS-Long preserves ultra-long memory by detachable KV cache mechanism shown in Fig~\ref{fig:stream}, resulting in much better performance.}
\vspace{-0.2cm}
\resizebox{0.99\textwidth}{!}{
\begin{tabular}{l|c|c|c|cccccccc}
\toprule
\textbf{Model} & \begin{tabular}[c]{@{}c@{}}\textbf{Num} \\ \textbf{Frame} \end{tabular} &\begin{tabular}[c]{@{}c@{}}\textbf{Token/} \\ \textbf{Frame}\end{tabular} &\begin{tabular}[c]{@{}c@{}}\textbf{Total Num}\\ \textbf{of Token}\end{tabular} & \textbf{LVBench} & \textbf{Video-MME} & \textbf{CG-Bench} & \textbf{Egoschema} & \textbf{MLVU} & \textbf{LongVideoBench} & \textbf{Avg} \\ \hline  \hline
VideoStreaming~\cite{qian2024streaming} & - & - & 256 & - & - & - & 44.1 & - & - & - \\
Qwen2-VL-online~\cite{wang2024qwen2} & - & - & 11520 & 39.8 & 59.4 & - & 64.0 & 62.9 & - & - \\
Flash-Vstream~\cite{zhang2025flash} & - & - & 11520 & 42.0 & 61.2 & - & 68.2 & 66.3 & - & - \\ \hline
\rowcolor{cyan!10} POINTS1.5-8B-online & 64 & 324 & 20736 & 41.7 & 59.3 & 33.1 & 59.3 & 64.7 & 53.5 & 51.9 \\ \hline
\rowcolor{orange!10} POINTS-Long & 248+8 & 8 & 3200 & 44.2 & 65.4 & 36.4 & 61.6 & 71.8 & 59.3 & 56.5 \\
\rowcolor{orange!10} POINTS-Long & 504+8 & 8 & 5248 & 46.0 & 65.0 & 35.5 & 59.7 & 70.3 & 59.2 & 56.0 \\ \hline
\rowcolor{orange!10} POINTS-Long & 248+8 & 16 & 5248 & 46.3 & 65.8 & 37.0 & 60.7 & 72.1 & 59.1 & 56.8 \\
\rowcolor{orange!10} POINTS-Long & 504+8 & 16 & 9344 & 48.6 & 64.9 & 35.6 & 58.4 & 71.2 & 58.8 & 56.3 \\
\bottomrule
\end{tabular}
}
\label{tab:stream}
\vspace{-0.1cm}
\end{table*}

\subsection{Implementation Details}
To balance between efficiency and fidelity, we compress single image into $n\in\{8,16,32\}$ tokens and set temporal $k=8$. For stage 1, we use the alignment data of POINTS1.5 and a subset of pretrain data. The newly introduced parameters were trained with learning rate 5e-5. For stage 2, we use high-quality data from pretrain and SFT stage, where the LLM parameters are unfrozen and jointly trained with learning rate 1e-5 (details are in supplementary material). The two-stage training process required approximately 25,000 H20 GPU hours. Note on Reproducibility: This work is primarily aimed at MLLM pre-training teams. The computational cost is highly dependent on the scale of proprietary training data and the size of the model.
\subsection{Evaluation \& Benchmarks}
\noindent \textbf{Fine-grained Image Benchmarks} We follow Opencompass~\cite{2023opencompass} image leaderboard, evaluating on MMBench~\cite{liu2024mmbench}, MathVista~\cite{lu2023mathvista}, HallusionBench~\cite{guan2024hallusionbench}, OCRBench~\cite{liu2024ocrbench}, AI2D~\cite{kembhavi2016diagram}, MMVet~\cite{yu2023mm}, MMStar~\cite{chen2024we}, MMMU~\cite{yue2024mmmu}.

\noindent \textbf{Video Benchmarks} We evaluate on a wide range of video benchmarks, including Opencompass video leaderboard: VideoMME~\cite{fu2025video}, Tempcompass~\cite{liu2024tempcompass}, MVBench~\cite{li2024mvbench}, MMBench-Video~\cite{fang2024mmbench}, MLVU~\cite{zhou2024mlvu}, LongVideoBench~\cite{wu2024longvideobench}, and other commonly used video benchmarks: MovieChat1K~\cite{song2024moviechat}, CG-Bench~\cite{chen2024cg}, EgoSchema~\cite{mangalam2023egoschema}, TemporalBench~\cite{cai2024temporalbench}, Activitynet-qa~\cite{caba2015activitynet}, LVBench~\cite{wang2025lvbench} and WorldSense~\cite{hong2025worldsense}. For Streaming understanding, we choose LongVideoBench, VideoMME, MLVU, LVBench, EgoSchema and CG-Bench.  We use VLMEvalKit~\cite{duan2024vlmevalkit} and lmms-eval~\cite{zhang2024lmms} for evaluation.

\begin{table*}[t]
\centering
\footnotesize
\caption{\textbf{Ablation Study.} We ablate the training design in Sec.~\ref{sec:overview}. All components are essential for obtaining the optimal result.}
\vspace{-0.2cm}
\resizebox{0.99\textwidth}{!}{
\begin{tabular}{c|c|c|c|cccccccc}
\toprule
\begin{tabular}[c]{@{}c@{}}\textbf{Num} \\ \textbf{Frame} \end{tabular} & \textbf{MLP} & \textbf{Temporal} & \textbf{Stage 2} & \textbf{MVBench} & \textbf{Video-MME} & \textbf{MMBench-Video} & \textbf{Tempcompass} & \textbf{MLVU} & \textbf{LongVideoBench} & \textbf{Avg} \\ \hline  \hline
64 & \textbf{$\times$} & \textbf{$\times$} & \textbf{$\times$} & 55.6 & 58.6 & 47.3 & 67.1 & 67.4 & 52.7 & 58.1 \\
64 & \textbf{\checkmark} & \textbf{\checkmark} & \textbf{$\times$} & 57.5 & 61.5 & 53.7 & 67.8 & 69.2 & 56.2 & 61.0 \\
64 & \textbf{\checkmark} & \textbf{$\times$} & \textbf{\checkmark} & 58.1 & 63.2 & 56.3 & 69.8 & 70.6 & 57.5 & 62.6 \\
64 & \textbf{\checkmark} & \textbf{\checkmark} & \textbf{\checkmark} & \textbf{59.4} & \textbf{63.5} & \textbf{58.0} & \textbf{69.9} & \textbf{71.9} & \textbf{58.2} & \textbf{63.5} \\ \hline
\bottomrule
\end{tabular}
}
\label{tab:ablation}
\vspace{-0.3cm}
\end{table*}

\begin{table}[t]
\centering
\footnotesize
\setlength{\tabcolsep}{3pt}
\caption{\textbf{Real-world Inference Speed-up.} We evaluate the speed-up of the LLM side prefilling and decoding using SGLang and Pytorch Profiler on H20. While ViT's cost grows linearly with the number of frames, LLM shows a quadratic increase in complexity.}
\vspace{-0.2cm}
\resizebox{0.48\textwidth}{!}{
\begin{tabular}{l|c|c|ccc}
\toprule
\textbf{Model} & \begin{tabular}[c]{@{}c@{}}\textbf{Num} \\ \textbf{Frame} \end{tabular} &\begin{tabular}[c]{@{}c@{}}\textbf{Token/} \\ \textbf{Frame}\end{tabular} & \begin{tabular}[c]{@{}c@{}}\textbf{FLOPs} \\ \textbf{(ViT+LLM)} \end{tabular} & \begin{tabular}[c]{@{}c@{}}\textbf{LLM Prefill} \\ \textbf{Latency (s)} \end{tabular} & \begin{tabular}[c]{@{}c@{}}\textbf{Generation} \\ \textbf{Throughput (token/s)}\end{tabular}\\ \hline  \hline
\rowcolor{cyan!10} POINTS1.5-8B & 128 & 144 & 99.7+456.7 & 3.23 & 311 \\
\rowcolor{cyan!10} POINTS1.5-8B & 256 & 144 & 199.4+1314.5 & 8.95 & 240 \\ \hline
\rowcolor{orange!10} POINTS-Long & 128 & 8 & 106.2+14.8 & 0.21 (6.5\%) & 1887 ($6.1\times$) \\
\rowcolor{orange!10} POINTS-Long & 256 & 8 & 208.5+30.9 & 0.41 (4.6\%) & 1494 ($6.2\times$) \\
\rowcolor{orange!10} POINTS-Long & 128 & 16 & 110.8+30.9 & 0.41 (12.7\%) & 1447 ($4.7\times$)\\
\rowcolor{orange!10} POINTS-Long & 256 & 16 & 221.6+66.8 & 0.70 (7.8\%) & 1124 ($4.7\times$) \\
\bottomrule
\end{tabular}
}
\label{tab:profiler}
\vspace{-0.4cm}
\end{table}

\subsection{Main Results}
\subsubsection{General Video Understanding}
\label{sec:general}
In Tab.~\ref{tab:opencompass-video} and ~\ref{tab:video-bench}, we compare POINTS-Long with base model POINTS1.5-8B-Instruct on a wide range of video benchmarks, under the same setting. As shown in Tab.~\ref{tab:opencompass-video}, with only 2.5\% to 10\% of the original tokens, our Standby Mode retains 97.7\% to 99.7\% of the full performance. Similar results are observed in Tab.~\ref{tab:video-bench} on more benchmarks.

This high level of performance retention, achieved through our native dual-mode training, significantly outperforms all prior visual token compression schemes, e.g. PruneVid~\cite{huang2024prunevid} retains only 96.9\% at a 10\% token ratio. In Tab.~\ref{tab:opencompass-video}, we report results using avg-pooling and low-resolution. Even with 4 times fewer tokens, POINTS-Long outperforms the baselines by a large margin (+4.3\%). Notably, when operating in Focus Mode, the model's performance is fully maintained (65.2 vs 65.0). This allows users to fully leverage the flexibility of our dual-mode system.

Furthermore, our model is designed for practical deployment: it requires no hyperparameter tuning, works out-of-the-box, and can be easily deployed in modern inference frameworks, making it ideal for industrial applications.


\subsubsection{Scalability of Frames in Inference}
\label{sec:frame}
We observe that for general MLLMs, video understanding performance stops increasing when exceeding certain number of frames (e.g. 64)~\cite{zohar2025apollo,jiang2025token}. We attribute this phenomenon to the LLM's long-range decay, stemming from inherent limitations in context length and their training data.

To validate the assumption, we evaluated our Standby Mode on long-video understanding benchmarks. As shown in Tab.~\ref{tab:frame}, the base model's performance doesn't improve much when scaling from 64 to 128 frames. In contrast, the Standby Mode of POINTS-Long shows continuously improving performance as the number of input frames increases. This allows it to achieve superior results on long-video tasks, all while using a much smaller token budget.

Note that POINTS1.5 was never trained on data over 128 frames. This zero-shot scalability is a remarkable property that only manifests in Standby Mode. We defer a detailed theoretical explanation for this phenomenon to future work.

\subsubsection{Fine-grained Image Understanding}
\label{sec:image}
A core design principle of POINTS-Long is the preservation of its fine-grained understanding capabilities. Unlike specialized video models~\cite{liu2025video,shu2024video,li2024videochat}, POINTS-Long completely retains the original model's fine-grained image understanding abilities through its Focus Mode. As shown in Tab.~\ref{tab:image}, POINTS-Long (Focus Mode) matches the base model's performance (69.7 vs 69.5), proving that our dual-mode training process is strictly beneficial and non-destructive to the model's core capabilities.

Furthermore, as an extra bonus, the learnable tokens can be leveraged to perform training-free visual token pruning. Specifically, by using the average attention weights from the learnable tokens in the final ViT layer to all visual tokens, we retain only the top $m$\% with the highest scores for the LLM (details in the supplementary material). This simple, training-free method yields impressive results. Compared to avg-pooling and other plug-and-play techniques~\cite{wang2025folder}, our attention-based pruning method achieves significantly better performance retention at the same compression ratios.

\subsubsection{Streaming Video Inference}
\label{sec:stream}
Streaming video understanding demands both fine-grained understanding of recent events and robust long-term memory. Standard MLLMs~\cite{bai2025qwen2,liu2024points1,wang2025internvl3} fail the latter: as new frames are prefilled, the context window is exhausted, forcing the earliest KV cache to be discarded. Such "sliding window" methods~\cite{xu2025streamingvlm} yield only short-term memory. Conversely, specialized streaming models~\cite{qian2024streaming,zhang2025flash} sacrifice the former, lacking critical fine-grained understanding.
  
POINTS-Long, however, is inherently well-suited for such a scenario via its dual-mode system. To validate the claim, we compare two setups. The baseline model is limited to a 64-frame sliding window, discarding older frames' KV cache. POINTS-Long, by contrast, activates its dual-mode strategy (Sec.~\ref{sec:inference}): the most recent 8 frames in Focus Mode (local window) and all preceding frames in Standby Mode (memory bank). We evaluate at the end of the video to test long-term recall. As shown in Tab.~\ref{tab:stream}, the baseline fails due to information loss, whereas POINTS-Long achieves superior performance by its high-quality memory.

\subsubsection{Ablation Study}

Tab.~\ref{tab:ablation} validates our key design choices. Duplicating the MLP layers enhances fitting capability, significantly boosting visual distillation. The temporal attention layer models temporal redundancy for more compact compression, further enhancing video understanding. Finally, our two-stage training is crucial: the second stage substantially improves Standby Mode while successfully preserving the Focus Mode's fine-grained understanding. Our final design, combining all components, yields the best performance.

\subsubsection{Inference Efficiency and Performance}
A primary motivator for POINTS-Long is computational efficiency. Prior research on visual token compression~\cite{chen2024image,ye2025voco,zhang2024sparsevlm} focus solely on algorithms, overlooking the practical deployment. Here, we analyze the acceleration benefits of Standby Mode from an infrastructure perspective. We divide MLLM into two components: ViT and LLM, which have very different computational and memory profiles.

\noindent \textbf{Disparate Workloads} The ViT-LLM computational balance is task-dependent. (1) For high-resolution images: For $\sim$10B models, compute is surprisingly comparable, as techniques like pixel-shuffle shortens the LLM's sequence. (2) For long videos: The bottleneck shifts to LLM. ViT compute scales linearly with frames, whereas the LLM's prefill scales quadratically, creating a dominant cost (Tab.~\ref{tab:profiler}).

\noindent \textbf{Distinct Compute Phases} The ViT encoding and LLM prefill phases are compute-intensive, while the LLM decode phase is I/O bound, as its speed is primarily limited by reading the KV cache. Infrastructure optimizations like continuous batching~\cite{kwon2023efficient} maximize throughput by batching parallel decode requests. The primary factor limiting this batch size is the available VRAM for the KV cache.


Based on these characteristics, our Standby Mode provides two crucial acceleration benefits:

\noindent \textbf{(1) Drastic Reduction in LLM Compute and Latency} As shown in Tab.~\ref{tab:profiler}, Standby Mode slashes the LLM's compute by 30-40$\times$. While ViT compute is not reduced, it can be overlapped with LLM, similar to "decoupled deployment" (PD) schemes~\cite{zhong2024distserve,wang2025internvl3}. This optimization is paramount, as the LLM's total time (prefill + decode) typically exceeds the ViT's encode time, making it the primary bottleneck.

\noindent \textbf{(2) Increased Generation Throughput via Batching} By reducing the visual sequence length, the KV cache footprint per sample becomes drastically smaller. This allows the inference system to batch significantly more concurrent decode requests within the same VRAM budget. This directly translates to a massive improvement (6.2$\times$) in overall generation throughput, a critical metric for production services.

We validated these claims using SGLang~\cite{zheng2024sglang} and PyTorch Profiler. Despite our implementation being preliminary and not fully optimized, the practical benefits are already substantial. As shown in Tab.~\ref{tab:profiler}, Standby Mode significantly reduces LLM prefill latency and boosts generation throughput. These advantages are especially pronounced in multi-frame video scenarios, confirming our approach's effectiveness for processing long visual sequences.

\section{Conclusion}
We introduce POINTS-Long, a novel dual-mode MLLM addressing the trade-off between fine-grained performance and computational efficiency. Inspired by human cognition, POINTS-Long operates in a high-fidelity "Focus Mode" and a highly compressed "Standby Mode". Our two-stage post-training strategy effectively integrates Standby Mode while fully preserving fine-grained abilities. POINTS-Long achieves state-of-the-art efficiency, retaining 97.7\%-99.7\% performance with only 1/40-1/10th visual tokens. Its dual-mode architecture also enables an efficient detachable KV cache for long-term streaming video understanding. Compatible with modern inference frameworks like SGLang, POINTS-Long offers a practical and powerful solution to the challenging trade-off in MLLM visual understanding.

\clearpage
{
    \small
    \bibliographystyle{ieeenat_fullname}
    \bibliography{main}
}

\clearpage
\setcounter{page}{1}

In this supplementary material, we first provide a comprehensive description of our base model, POINTS1.5-8B-Instruct. Subsequently, we elaborate on the architectural details and training protocols of POINTS-Long. Finally, we present additional ablation studies and visualizations.

\section{Details about POINTS1.5-8B-Instruct} \label{sec:points}
\subsection{Model Architecture}
\begin{figure*}
  \centering
  \includegraphics[width=1\textwidth]{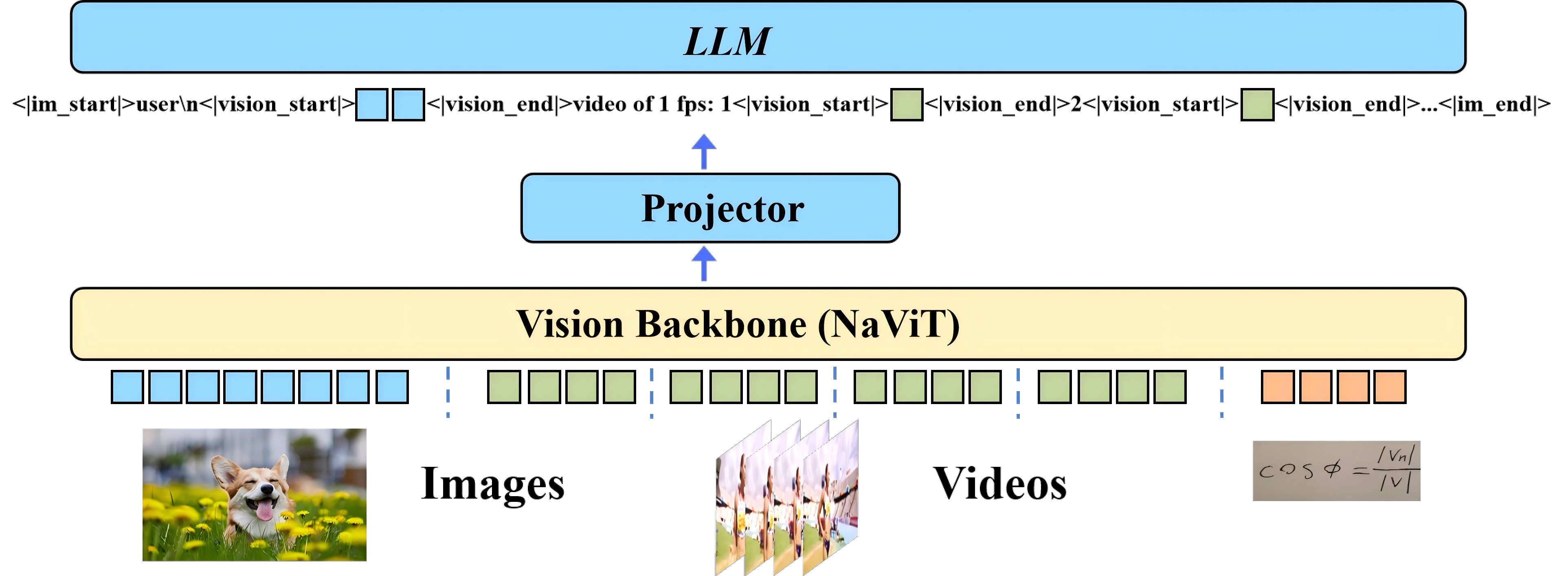}
  \caption{\textbf{POINTS1.5-8B-Instruct Architecture.} POINTS1.5-8B consists of a native-resolution image encoder (initialized from Qwen2-VL-ViT), a pixel-shuffle projector reducing the token count by a factor of 4, and an LLM initialized from Qwen3-8B-Base. The architecture employs 1D RoPE for the LLM and 2D RoPE for the ViT.}
  \label{fig:point}
\end{figure*}

The POINTS~\cite{liu2024points1} series is a family of advanced multimodal large language models (MLLMs) that was first released in September 2024. The POINTS1.5-8B-Instruct model employed in this work is an enhanced iteration of POINTS1.5~\cite{liu2024points1} (Fig.~\ref{fig:point}). It is initialized from Qwen3-8B-Base~\cite{yang2025qwen3} and Qwen2-VL-ViT~\cite{wang2024qwen2}. The model applies 1D RoPE~\cite{su2024roformer} for visual tokens within the LLM backbone and 2D RoPE within the ViT image encoder. Furthermore, the intermediate projector utilizes a pixel-shuffle operation to reduce the visual sequence length by a factor of 4.
\subsection{Model Training Dataset}
\label{sec:dataset}
POINTS1.5-8B-Instruct underwent comprehensive multimodal training, organized into four distinct stages:

\noindent \textbf{Visual-textual Alignment} In this initial phase, the parameters of both the Vision Transformer (ViT) and the Large Language Model (LLM) remained frozen, with training optimized solely on the alignment projector. We utilized Laion-5B~\cite{schuhmann2022laion} as seed data, which was subsequently processed using CapFusion~\cite{yu2024capsfusion} for recaptioning and perplexity filtering~\cite{liu2024points} for quality control. For this stage, we employed a sequence length of 8192.

\noindent \textbf{Multimodal Continue Pre-training} To construct our image-text pre-training dataset, we sourced raw PDFs from the CC-MAIN-2021-31-PDF-UNTRUNCATED\footnote{https://digitalcorpora.org/corpora/file-corpora/cc-main-2021-31-pdf-untruncated/} dataset, retaining only Chinese and English documents. Our processing pipeline utilized PaddleOCR~\cite{cui2025paddleocr} for image extraction and the POINTS-Reader~\cite{liu2025points} document OCR model for text extraction. For each document, we concatenated the extracted images (placed at the beginning) with the corresponding text and page format. This process yielded a pre-training corpus containing approximately 400 billion tokens. Analogous to LLM pre-training, this stage utilizes massive unlabeled web data to expose the model to broad world knowledge.

\noindent \textbf{Multimodal Decay} Following pre-training, we initiated a Decay stage designed to bolster the MLLM's performance across a spectrum of capabilities, including grounding, OCR, GUI navigation, reasoning, video understanding, and text-based CoT.

To achieve this, we constructed specialized training data from diverse sources, including open-source datasets such as Wukong~\cite{gu2022wukong}, Object365~\cite{shao2019objects365}, and Koala-36M~\cite{wang2025koala}, alongside proprietary in-house data. Training in this stage was conducted in two steps: first, we focused on fine-grained image understanding with a context length of 8k. Subsequently, we expanded the context length to 32k, incorporating data for complex video understanding tasks (e.g., dense captioning and temporal grounding) and long-context, text-only CoT data.

\noindent \textbf{Multimodal Supervised Instruction Tuning} The Multimodal Supervised Fine-Tuning (SFT) stage is designed to utilize high-quality data to teach the model to follow instructions and align with human preferences.

In this phase, we utilize a large volume of high-quality image-text and video QA data. For the video domain, in addition to proprietary in-house data, we primarily leverage open-source datasets, including FineVideo~\cite{Farré2024FineVideo}, Vript~\cite{yang2024vript}, ShareGPT4Video~\cite{chen2024sharegpt4video}, OpenVid-1M~\cite{nan2024openvid}, VideoUFO~\cite{wang2025videoufo}, CinePile~\cite{rawal2024cinepile}, VideoChat2IT~\cite{li2023videochat}, LLaVA-Hound~\cite{zhang2025direct}, LLaVA-Video-178K~\cite{zhang2024video}, and Ego4D~\cite{grauman2022ego4d}. We conduct training with a 32K context length in this stage. For video preprocessing, we split ultra-long videos into shorter segments and sample frames at 1 fps. Due to sequence length constraints, we set the maximum frame limit to 128; videos exceeding this limit are uniformly downsampled on temporal dimension.

\noindent \textbf{Multimodal Post-training} We apply RFT (Rejection Sampling Fine-Tuning) and RL (Reinforcement Learning) to enhance the model's reasoning and cognitive capabilities. For RFT, we utilize open-source synthetic reasoning datasets such as AM-DeepSeek-R1-Distilled-1.4M~\cite{zhao20251}, Reason-RFT~\cite{tan2025reason}, and VisualWebInstruct~\cite{jia2025visualwebinstruct}, covering a wide range of disciplines. For RL, we train on a diverse range of tasks, including STEM (e.g., mathematics, physics, chemistry), puzzle solving, and OCR-based reasoning.

\subsection{Training Recipe} 
We conduct training using our in-house framework, which is analogous to Megatron~\cite{shoeybi2019megatron}. We set the Tensor Parallel (TP) degree to 2 during the Alignment stage and 4 for all subsequent stages. During the long-context training phase, we enable sequence parallelism and utilize activation checkpointing to minimize memory overhead. We employ a "pack-to-pack" (or sample packing) training strategy, utilizing a learning rate of 3e-4 for the Alignment stage and 5e-5 for all other phases.

\subsection{Chat Template} 
We adhere to the standard chat template of Qwen2.5-VL~\cite{bai2025qwen2}, with the primary distinction lying in the representation of video inputs. Instead of treating the video as a monolithic entity, we enclose each input frame within \textit{\texttt{<\textbar vision\_start\textbar>} \texttt{<\textbar vision\_end\textbar>}} tags. To enable the model to explicitly perceive temporal information, we prepend a metadata string to the video input: \texttt{Video of x fps:}. This prefix identifies the modality and specifies the framerate. Furthermore, we interleave textual timestamps between video frames. To conserve token usage, these timestamps are inserted directly as numerical values representing seconds (e.g., \texttt{1<frame1>2.5<frame2>4<frame3>}).

\section{Details about POINTS-Long} \label{sec:points}
In the supplementary material, we provide more details about POINTS-Long.
\subsection{POINTS-Long Architecture}
POINTS-Long is built upon the POINTS1.5-8B-Instruct architecture. As illustrated in Fig. 2 of the main paper, the primary modification involves the vision backbone, where $n$ additional learnable tokens—termed "standby tokens"—are concatenated with the original patchified visual sequence. Within each layer, duplicated MLPs are introduced to process these standby tokens independently. Furthermore, a temporal modeling attention block is inserted into the final 5 layers of the ViT to encode standby tokens across 8 adjacent frames. Crucially, the attention mechanism in this temporal block is causal, enabling efficient processing of streaming inputs without the need for re-computation. Unlike full attention, which necessitates simultaneous access to a window of 8 frames during the forward pass—an approach ill-suited for frame-by-frame streaming—causal attention allows the model to simply cache the standby representations of the preceding 7 frames. This results in negligible memory overhead while significantly enhancing the model's capability to handle streaming scenarios.

To maintain architectural consistency, we apply the same duplication strategy to the projection layer. It is important to clarify the notation regarding $n$: in our experimental tables, the reported token count refers to the final input tokens to the LLM. Since the projection layer employs a pixel-shuffle operation that aggregates 4 neighboring tokens into 1, the number of learnable standby tokens initialized in the ViT is 4 times the final token count in the LLM. For instance, in Tab. 1 in the main paper, a "Num/Frame" of 8 corresponds to an initialization of $n=32$ standby tokens in the vision backbone.

Here we express this encoding process in a mathematical way. The input image $I_q$ is transformed into a patchified visual sequence with $o$ as sequence length: $Z_{q0}=\{z_{q01}, ..., z_{q0o}\} \in \mathbb{R}^{o\times d}$ by patch embedding layer (the 3 subscripts represent frame index, layer index, and sequence index, respectively). We initialize $n$ learnable tokens $L_{q0}=\{l_{q01}, ..., l_{q0n}\} \in \mathbb{R}^{n\times d}$ and prepend them to the original sequence $\{L_{q0},Z_{q0}\}=\{l_{q01},...l_{q0n}, z_{q01}, ..., z_{q0o}\}$, where normally $o\gg n$. The two parallel sequences share the same attention block:
\begin{equation}
    \{L'_{qi},Z'_{qi}\} = \text{Attention\_Block}_{i}(\{L_{qi},Z_{qi}\}),
\end{equation}
where $i$ is the layer/block index.
The resultant sequences are processed by different MLPs:
\begin{equation}
    \{L_{q(i+1)},Z_{q(i+1)}\} = \{\text{MLP}_{Li}(L'_{qi}),\text{MLP}_{Zi}(Z'_{qi})\}.
\end{equation}
Note that the parameter of $\text{MLP}_{Li}$ is initialized by $\text{MLP}_{Zi}$. In the last 5 blocks, we add one temporal attention between attention and MLP, taking only the learnable tokens of the adjacent 8 frames:
\begin{equation}
\begin{split}
    &\{L''_{(q-w)i},...,L''_{qi},...,L''_{(q+v)i}\} = \\
         &\text{Attenion\_T}_{i}(\{L'_{(q-w)i},...,L'_{qi},...,L'_{(q+v)i}\}),
\end{split}
\end{equation}
where $w+v \leq 8$, depending on the position of current input image/frame $I_q$. For image understanding, the input is $L'_{qi}$ only, and for video inputs, we group the neighboring 8 frames without overlap. Since we use pack-to-pack parallel computing technique, the temporal attention only needs to be calculated once per 8 frames. With temporal modeling, the subsequent MLP layer becomes:
\begin{equation}
    \{L_{q(i+1)},Z_{q(i+1)}\} = \{\text{MLP}_{Li}(L''_{qi}),\text{MLP}_{Zi}(Z'_{qi})\}.
\end{equation}

For the projection layer, we also apply the same parallel encoding strategy. We note $\{z_1, ..., z_o\}_q$ the resultant original visual sequence for each image $q$ and $\{l_1,...,l_n\}_q$ the learnable standby tokens, after being encoded by ViT and projector. During the two-stage training process, we activates different modes. In stage 1, we only pass the learnable standby tokens to LLM:
\begin{equation}
    \text{Loss}=LLM(\{l_1, ..., l_n\}_q \forall q \in Q,\text{Text}), \\
\end{equation}
where $Q$ is the image/frame set in the sample. In stage 2, we apply the 2-forward training strategy:
\begin{equation}
\begin{split}
    &\text{Loss}_1=LLM(\{l_1, ..., l_n\}_q \forall q \in Q,\text{Text}), \\
    &\text{Loss}_2=LLM(\{l_1, ..., l_n, z_1, ..., z_o\}_q \forall q \in Q,\text{Text}), \\
    & \text{Loss} = \frac{1}{2}(\text{Loss}_1 + \text{Loss}_2)
\end{split}
\end{equation}

\begin{table*}[t]
\centering
\footnotesize
\setlength{\tabcolsep}{7pt}
\caption{\textbf{Performance of Different Inference Mode.} In standard focus mode, we concatenate the learnable standby tokens with the original visual tokens and pass to LLM. Nevertheless, it makes no big difference to inference with only the original visual tokens. *ori-seq means original sequence without standby tokens.}
\vspace{-0.2cm}
\resizebox{0.99\textwidth}{!}{
\begin{tabular}{l|cccccccccc}
\toprule
\textbf{Model} & \textbf{MMBench} & \textbf{MMStar} & \textbf{MMMU\_val} & \textbf{MathVista} & \textbf{OCRBench} & \textbf{AI2D} & \textbf{HallusionBench} & \textbf{MMVet} & \textbf{Avg} \\ \hline  \hline
\rowcolor{cyan!10} POINTS1.5-8B (baseline) & 81.9 & 65.7 & 53.2 & 70.9 & 85.8 & 83.9 & 50.1 & 64.7 & 69.5 \\
\rowcolor{orange!10} POINTS-Long (focus) & 82.1 & 66.1 & 53.7 & 70.6 & 85.5 & 84.2 & 48.3 & 66.7 & 69.7 \\
\rowcolor{orange!10} POINTS-Long (ori-seq) & 82.1 & 65.8 & 53.0 & 69.7 & 85.0 & 83.8 & 48.0 & 67.4 & 69.4 \\
\bottomrule
\end{tabular}
}
\label{tab:focus}
\end{table*}

\begin{table*}[t]
\centering
\footnotesize
\setlength{\tabcolsep}{4pt}
\caption{\textbf{Learning Rate \& Model Performance.} We train the model under different learning rates (1e-5, 2e-5, 5e-5) in stage 2. Performance differences were minimal, proving the training scheme's robustness.}
\vspace{-0.2cm}
\resizebox{0.99\textwidth}{!}{
\begin{tabular}{l|c|c|c|ccccccl}
\toprule
\textbf{Model} & \begin{tabular}[c]{@{}c@{}} \textbf{Num} \\ \textbf{Frame} \end{tabular} &\begin{tabular}[c]{@{}c@{}} \textbf{Token/} \\ \textbf{Frame}\end{tabular} &\begin{tabular}[c]{@{}c@{}}\textbf{Learning}\\ \textbf{Rate}\end{tabular} & \textbf{MVBench} & \textbf{Video-MME} & \textbf{MMBench-Video} & \textbf{Tempcompass} & \textbf{MLVU} & \textbf{LongVideoBench} & \textbf{Avg} \\ \hline  \hline
 POINTS-Long (standby) & 64 & 16 & 1e-5 & 59.7 & 65.0 & 59.3 & 69.1 & 71.7 & 58.9 & \textbf{63.9} \\
 POINTS-Long (standby) & 64 & 16 & 2e-5 & 58.3 & 64.9 & 59.3 & 69.3 & 71.8 & 58.2 & 63.6 \\
 POINTS-Long (standby) & 64 & 16 & 5e-5 & 59.7 & 64.9 & 60.0 & 69.4 & 70.3 & 58.4 & 63.8 \\ \hline
 POINTS-Long (standby) & 64 & 32 & 1e-5 & 60.8 & 65.7 & 60.9 & 70.3 & 71.6 & 59.5 & \textbf{64.8} \\
 POINTS-Long (focus) & 64 & 32 & 1e-5 & 61.0 & 66.1 & 60.3 & 71.3 & 73.2 & 59.4 & 65.2 \\
 POINTS-Long (standby) & 64 & 32 & 2e-5 & 61.3 & 65.9 & 60.3 & 70.6 & 70.8 & 59.5 & 64.7 \\
 POINTS-Long (focus) & 64 & 32 & 2e-5 & 62.1 & 66.1 & 61.3 & 71.4 & 72.5 & 58.8 & \textbf{65.4} \\
 POINTS-Long (standby) & 64 & 32 & 5e-5 & 58.8 & 65.8 & 61.3 & 71.3 & 71.1 & 59.4 & 64.6 \\
 POINTS-Long (focus) & 64 & 32 & 5e-5 & 61.2 & 67.0 & 61.3 & 71.1 & 72.2 & 59.5 & 65.4 \\
\bottomrule
\end{tabular}
}
\vspace{-0.2cm}
\label{tab:lr}
\end{table*}

\begin{table*}[t]
\centering
\footnotesize
\setlength{\tabcolsep}{4pt}
\caption{\textbf{Training Data \& Model Performance.} We train the model using different amount of data in stage 2. By adding more high-quality data in stage 2 (85\%-100\%), we witness a steady improvement in performance.}
\vspace{-0.3cm}
\resizebox{0.99\textwidth}{!}{
\begin{tabular}{l|c|c|c|ccccccl}
\toprule
\textbf{Model} & \begin{tabular}[c]{@{}c@{}} \textbf{Num} \\ \textbf{Frame} \end{tabular} &\begin{tabular}[c]{@{}c@{}} \textbf{Token/} \\ \textbf{Frame}\end{tabular} & \begin{tabular}[c]{@{}c@{}} \textbf{Training} \\ \textbf{Data}\end{tabular} & \textbf{MVBench} & \textbf{Video-MME} & \textbf{MMBench-Video} & \textbf{Tempcompass} & \textbf{MLVU} & \textbf{LongVideoBench} & \textbf{Avg} \\ \hline  \hline
 POINTS-Long (standby) & 64 & 16 & Standard & 59.7 & 65.0 & 59.3 & 69.1 & 71.7 & 58.9 & \textbf{63.9} \\
 POINTS-Long (standby) & 64 & 16 & Reduced & 59.3 & 64.0 & 59.0 & 68.9 & 71.3 & 59.2 & 63.6 \\ \hline
 POINTS-Long (standby) & 64 & 32 & Standard & 60.8 & 65.7 & 60.9 & 70.3 & 71.6 & 59.5 & \textbf{64.8} \\
 POINTS-Long (standby) & 64 & 32 & Reduced & 60.7 & 65.1 & 59.3 & 69.9 & 70.8 & 60.9 & 64.3 \\
\bottomrule
\end{tabular}
}
\vspace{-0.1cm}
\label{tab:dataset}
\end{table*}

\begin{table*}[t]
\centering
\footnotesize
\setlength{\tabcolsep}{5pt}
\caption{\textbf{Comparison with Visual Token Reduction Methods.} Under the same setting (or even using fewer tokens), POINTS-Long exceeds previous visual token reduction methods by a large margin. It's a natural result since the standby mode is carefully trained as a native inference mode. }
\vspace{-0.3cm}
\resizebox{0.99\textwidth}{!}{
\begin{tabular}{l|c|c|c|ccccccl}
\toprule
\textbf{Model} & \begin{tabular}[c]{@{}c@{}} \textbf{Num} \\ \textbf{Frame} \end{tabular} &\begin{tabular}[c]{@{}c@{}} \textbf{Token/} \\ \textbf{Frame}\end{tabular} &\begin{tabular}[c]{@{}c@{}}\textbf{Total Num}\\ \textbf{of Token}\end{tabular} & \textbf{MVBench} & \textbf{Video-MME} & \textbf{MMBench-Video} & \textbf{Tempcompass} & \textbf{MLVU} & \textbf{LongVideoBench} & \textbf{Avg} \\ \hline  \hline
\rowcolor{cyan!10} POINTS1.5-8B (baseline) & 64 & 324 & $\approx$ 20K & 60.3 & 66.1 & 61.0 & 71.1 & 72.0 & 59.8 & 65.0 \\
POINTS1.5-8B (low-resolution) & 64 & 32 & 2048 (10\%) & 54.9 & 61.2 & 51.0 & 67.1 & 67.3 & 53.9 & 59.2 (91.1\%) \\
POINTS1.5-8B (pooling) & 64 & 32 & 2048 (10\%) & 54.9 & 55.4 & 43.0 & 66.6 & 67.1 & 54.5 & 56.9 (87.5\%) \\
POINTS1.5-8B (+VisionZip~\cite{yang2025visionzip}) & 64 & 32 & 2048 (10\%) & 56.7 & 60.0 & 51.7 & 66.6 & 65.9 & 54.9 & 59.3 (91.2\%) \\
POINTS1.5-8B (+Dycoke~\cite{tao2025dycoke}) & 64 & 32 & 2048 (10\%) & 56.1 & 62.5 & 55.7 & 67.5 & 67.2 & 55.6 & 60.8 (93.5\%) \\
POINTS1.5-8B (+PruneVID~\cite{huang2024prunevid}) & 64 & 32 & 2048 (10\%) & 57.8 & 62.0 & 55.3 & 69.3 & 67.5 & 56.5 & 61.4 (94.4\%)\\
POINTS1.5-8B (+FastVID~\cite{shen2025fastvid}) & 64 & 32 & 2048 (10\%) & 54.9 & 63.9 & 56.0 & 68.3 & 70.1 & 54.5 & 62.7 (96.5\%) \\
\rowcolor{orange!10} POINTS-Long (standby) & 64 & 8 & 512 (2.5\%) & 59.4 & 63.5 & 58.0 & 69.9 & 71.9 & 58.2 & 63.5 (97.7\%) \\
\rowcolor{orange!10} POINTS-Long (standby) & 64 & 16 & 1024 (5\%) & 59.7 & 65.0 & 59.3 & 69.1 & 71.7 & 58.9 & 63.9 (98.3\%) \\
\rowcolor{orange!10} POINTS-Long (standby) & 64 & 32 & 2048 (10\%) & 60.8 & 65.7 & 60.9 & 70.3 & 71.6 & 59.5 & 64.8 (99.7\%) \\
\bottomrule
\end{tabular}
}
\vspace{-0.1cm}
\label{tab:opencompass-video}
\end{table*}

\begin{table*}[t]
\centering
\footnotesize
\setlength{\tabcolsep}{4pt}
\caption{\textbf{Model Soup Performance.} We apply model soup (model merge) to two models trained by different learning rates. The model's performance can further boost in this way.}
\vspace{-0.3cm}
\resizebox{0.99\textwidth}{!}{
\begin{tabular}{l|c|c|c|ccccccl}
\toprule
\textbf{Model} & \begin{tabular}[c]{@{}c@{}} \textbf{Num} \\ \textbf{Frame} \end{tabular} &\begin{tabular}[c]{@{}c@{}} \textbf{Token/} \\ \textbf{Frame}\end{tabular} &\begin{tabular}[c]{@{}c@{}}\textbf{Total Num}\\ \textbf{of Token}\end{tabular} & \textbf{CG-Bench} & \textbf{Video-MME} & \textbf{MMBench-Video} & \textbf{Tempcompass} & \textbf{MLVU} & \textbf{LongVideoBench} & \textbf{Avg} \\ \hline  \hline
\rowcolor{cyan!10} POINTS1.5-8B (baseline) & 64 & 324 & $\approx$ 20K & 36.7 & 66.1 & 61.0 & 71.1 & 72.0 & 59.8 & 61.1 \\
\rowcolor{orange!10} POINTS-Long (standby) & 64 & 16 & 1024 (5\%) & 34.6 & 65.0 & 59.3 & 69.1 & 71.7 & 58.9 & 59.8 \\
\rowcolor{orange!10} POINTS-Long (model soup) & 64 & 16 & 1024 (5\%) & 35.2 & 65.9 & 60.3 & 69.9 & 71.6 & 58.1 & 60.2 (+0.4) \\
\rowcolor{orange!10} POINTS-Long (standby) & 128 & 16 & 2048 (10\%) & 36.2 & 66.4 & 61.0 & 69.6 & 72.7 & 60.3 & 61.0 \\
\rowcolor{orange!10} POINTS-Long (model soup) & 128 & 16 & 2048 (10\%) & 36.7 & 66.7 & 61.3 & 70.4 & 72.0 & 60.7 & 61.3 (+0.3) \\
\rowcolor{orange!10} POINTS-Long (standby) & 64 & 32 & 2048 (10\%) & 35.7 & 65.7 & 60.9 & 70.3 & 71.6 & 59.5 & 60.6 \\
\rowcolor{orange!10} POINTS-Long (model soup) & 64 & 32 & 2048 (10\%) & 36.5 & 65.8 & 61.0 & 70.9 & 72.1 & 59.8 & 61.0 (+0.4) \\
\rowcolor{orange!10} POINTS-Long (standby) & 128 & 32 & 4096 (20\%) & 37.3 & 66.9 & 62.0 & 70.1 & 72.5 & 60.4 & 61.5 \\
\rowcolor{orange!10} POINTS-Long (model soup) & 128 & 32 & 4096 (20\%) & 37.5 & 66.6 & 63.3 & 70.8 & 73.8 & 61.2 & 62.2 (+0.7) \\
\bottomrule
\end{tabular}
}
\vspace{-0.3cm}
\label{tab:soup}
\end{table*}

\subsection{Training Dataset}
The training of POINTS-Long is conducted in two distinct stages.

\noindent \textbf{Stage 1: Visual Distillation and Alignment.} In this phase, all parameters of the original architecture—including the LLM backbone—remain frozen. Optimization is restricted exclusively to the newly introduced learnable tokens, the duplicated MLPs, and the projection layer. The objective is to enable these ``standby tokens'' to effectively aggregate and distill visual information from the original sequence, a process analogous to the alignment phase in MLLM training. To achieve this, we utilize the complete alignment dataset alongside a subset of data from the multimodal decay stage (detailed in Sec.~\ref{sec:dataset}) to ensure robust visual distillation. Since all parameters governing the original inference path remain frozen, the model's baseline performance remains strictly preserved during this stage.

\noindent \textbf{Stage 2: LLM Mode Adaptation.} In the second stage, we fine-tune the LLM using a reduced learning rate. We employ a dual-path forward strategy: computing the average loss derived from both Standby and Focus forward passes before backpropagating the gradients. This mechanism enables the LLM to adapt simultaneously to both inference modes. For this stage, we incorporate a high-quality subset of the multimodal decay data alongside the full Supervised Fine-Tuning (SFT) dataset. Notably, all training data employed across both stages is derived exclusively from the training set of the baseline model, POINTS1.5-8B. No external data is introduced, thereby ensuring a fair comparison.


\subsection{Dual-Mode Inference}
Here, we detail the inference protocols for standby mode and focus mode. When operating in standby mode, we feed only the compressed, short learnable token sequence to the LLM as visual input for inference. Conversely, in focus mode, the entire sequence—comprising both the learnable tokens and the original visual tokens—is passed to the LLM. Formally, for focus mode, we pass $\{l_1, ..., l_n, z_1, ..., z_o\}_q$ to LLM. While for standby mode, we only pass $\{l_1, ..., l_n\}_q$. Formally, we can express the inference of the two modes as follows: 
\begin{equation}
\begin{split}
    &\text{Standby} : \text{Output}=\text{LLM}(\{l_1, ..., l_n\}_q,\text{Text}), \\
    &\text{Focus}: \text{Output}=\text{LLM}(\{l_1, ..., l_n, z_1, ..., z_o\}_q,\text{Text})
\end{split}
\end{equation}

In practice, we could use only the original visual sequence (without the learnable tokens) for inference. However, including the learnable tokens provides a significant advantage for streaming visual inference: we can leverage the "detachable KV cache" technique (described in the main paper Sec.~3.4) and avoid re-computation. Given that the learnable token sequence is significantly shorter than the original sequence, this method has a negligible impact on accuracy and computational overhead, as we show in Tab.~\ref{tab:focus}.

\subsection{Training-free Token Pruning}
Following the two-stage training, the standby tokens have effectively absorbed critical visual information from the original sequence. Previous research has indicated that the attention distribution of learnable global representation tokens correlates strongly with the most salient information. This implies that we can leverage the attention distribution of the standby tokens relative to other tokens to identify the most important visual tokens within the long sequence.

This enables us to perform a training-free pruning of visual tokens based directly on this distribution. Specifically, within the final layer of the Vision Transformer (ViT), we calculate the mean attention score for each token in the original visual sequence relative to the standby tokens. These scores are then sorted, and we retain only the top m\% of tokens to be fed into the LLM.

It is important to note that the pixel-shuffle operation performs a projection on adjacent groups of four tokens. To avoid disrupting this projection, our compression method treats these four-token groups as atomic units. We calculate a group attention score (by averaging the attention of the four constituent tokens), ensuring that we prune at the granularity of these post-pixel-shuffle units. As demonstrated in Tab.~4 of the main paper, this straightforward approach outperforms other token compression methods.

\section{Details on Evaluation}
In this section, we explain in detail our evaluation metric.
\subsection{Evaluation Benchmark}
\noindent \textbf{Fine-grained Image Benchmarks} We leverage Opencompass~\cite{2023opencompass} image benchmark for evaluation, including MMBench~\cite{liu2024mmbench}, MathVista~\cite{lu2023mathvista}, HallusionBench~\cite{guan2024hallusionbench}, OCRBench~\cite{liu2024ocrbench}, AI2D~\cite{kembhavi2016diagram}, MMVet~\cite{yu2023mm}, MMStar~\cite{chen2024we}, MMMU~\cite{yue2024mmmu}. Note that MMMU is evaluated on validation set, MMBench is the average of MMBench\_test\_EN and MMBench\_test\_CN. We use VLMEvalKit~\cite{duan2024vlmevalkit} for all the image evaluation.

\noindent \textbf{Video Benchmarks} We evaluate on a wide range of video benchmarks, including Opencompass video leaderboard: VideoMME~\cite{fu2025video}, Tempcompass~\cite{liu2024tempcompass}, MVBench~\cite{li2024mvbench}, MMBench-Video~\cite{fang2024mmbench}, MLVU~\cite{zhou2024mlvu}, LongVideoBench~\cite{wu2024longvideobench}, and other commonly used video benchmarks: MovieChat1K~\cite{song2024moviechat}, CG-Bench~\cite{chen2024cg}, EgoSchema~\cite{mangalam2023egoschema}, TemporalBench~\cite{cai2024temporalbench}, Activitynet-qa~\cite{caba2015activitynet}, LVBench~\cite{wang2025lvbench} and WorldSense~\cite{hong2025worldsense}. We use lmms-eval~\cite{zhang2024lmms} to evaluate LVBench, WorldSense, EgoSchema, TemporalBench and Activitynet-qa, while for the rest we use VLMEvalKit~\cite{duan2024vlmevalkit}. Note that we evaluate CG-Bench on its long accuracy and MLVU on M-Avg.

\subsection{Streaming Video Evaluation}
Streaming video understanding is an increasingly critical application for large models. Our POINTS-Long model is specifically designed and optimized for this scenario, achieving a long-term visual memory bank by leveraging a detachable KV cache and dual-mode cooperation.

Our evaluation methodology mimics a real-world streaming scenario. For the baseline model's inference, we uniformly sample 256 frames for prefilling. Once its 64-frame context limit is exceeded, the preceding frame's KV cache is discarded.

For POINTS-Long, we uniformly sample either 256 or 512 frames. The most recent 8 frames are prefilled using focus mode. As soon as this 8-frame local window limit is surpassed, we follow the procedure illustrated in Fig.~3 of the main paper: the standby tokens of previous frames about to be discarded are detached and integrated into the memory bank, while the rest are dropped. While the precise system implementation for this detachment is complex, it can be simplified by just re-prefilling the standby tokens. This alternative method yields nearly identical results with a negligible increase in computational overhead.

It is important to note that while POINTS-Long substantially extends the visual memory capacity, high-FPS long videos may still surpass the context length limit. For this unavoidable forgetting, we recommend employing an external database, such as M3-agent~\cite{long2025seeing}. Since this component is outside the scope of the core POINTS-Long solution, we do not detail a specific implementation. Our evaluation employs a fixed number of frames specifically to measure the performance gains within the POINTS-Long memory capacity; performance on content exceeding this range is expected to be no different from the baseline model.

\subsection{Efficiency Benchmarking}
In Sec. 4.3.6 of the main paper, we provide a detailed analysis of the advantages of POINTS-Long for industrial-grade deployment. We emphasize that POINTS-Long can significantly accelerate inference in two key ways: 

\noindent \textbf{Substantial Reduction in LLM Prefill Time:} POINTS-Long significantly reduce the visual sequence length, thus speed up the LLM prefill phase. Benchmarks using SGLang measured a 10-20x decrease in LLM prefill latency.

\noindent \textbf{Increased Decode Throughput:} During the LLM decode stage, the historical visual sequence is drastically shortened. This allows us to parallelize significantly more decode requests under the same KV cache budget. Because decoding is an I/O-intensive operation, the number of parallel requests is almost directly proportional to the throughput. Even with our relatively naive implementation, we achieved a 6.2x increase in generation throughput.

For our benchmarks, we used identical samples (from VideoMME) and precisely measured LLM prefill latency using SGLang and the PyTorch profiler. To test throughput, we optimized SGLang's asynchronous visual input CPU preprocessing by using multiprocessing for frame handling, thereby increasing request parallelism. With mem-fraction-static=0.65, the baseline model using 256 frames could only decode approximately 8 requests in parallel. In contrast, POINTS-Long was able to decode over 70 requests in parallel. (It is worth noting that this number was constrained by our system's CPU performance and machine bandwidth, suggesting that the optimal parallel capacity can be higher.)
\section{More Experiment}
\subsection{Ablation}
In addition to the ablation study on the parameter module presented in the main paper, we conduct supplementary experiments regarding training data size and learning rate.

\noindent \textbf{Ablation on Learning Rate} 
As shown in Tab.~\ref{tab:lr}, we evaluated the model performance using varying learning rates (1e-5, 2e-5, 5e-5) on LLM at stage 2. The results on video benchmarks indicate minimal performance variance across Standby and Focus inference modes, thereby validating the robustness of our two-stage training strategy. Consequently, to preserve the model's general capabilities and minimize weight shifts, we use the smaller learning rate for stage 2.

\noindent \textbf{Ablation on Training Data} In Tab.~\ref{tab:dataset}, we demonstrate the effect of data scaling during Stage 2. Increasing the amount of high-quality image-text and video data yields consistent performance improvements. This validates the criticality of data scale and indicates promising scalability towards larger architectures and more extensive datasets.

\subsection{Comparison with Visual Reduction Methods}
Recent works have extensively explored visual token compression, particularly for video understanding~\cite{tao2025dycoke,chen2024image,yang2025visionzip,wang2025folder,huang2024prunevid,shen2025fastvid}. While most existing approaches are training-free—offering high compatibility—they suffer from severe performance degradation at high compression ratios (as discussed in our Introduction). POINTS-Long addresses this bottleneck via native training, effectively embedding the high-compression 'Standby' mode as a native inference mechanism. As shown in Table 6, this strategy allows POINTS-Long to significantly outperform previous methods at the same compression ratio (99.7\% vs. 96.5\%). Remarkably, even with 4 times fewer tokens, our model still achieves superior performance (97.7\%). This native training paradigm maximizes model potential and represents the future architectural direction for MLLMs. Note that for all comparison methods, we re-implement on POINTS1.5-8B-Instruct, using only their optimization before LLM.

\subsection{Model Soup Enhancement}
In POINTS1.5~\cite{liu2024points1}, we employ the Model Soup technique~\cite{wortsman2022model} to enhance performance. Model Soup involves averaging the weights of multiple fine-tuned models—often trained with different hyperparameters or data—to improve generalization without incurring additional inference costs. specifically, we performed simple parameter averaging on two model checkpoints trained with distinct learning rates. We observed a consistent and notable performance gain across benchmarks (ranging from +0.3 to +0.7). This indicates that the model has not yet reached its performance upper bound and has further capacity for optimization.

\begin{figure*}[h]
  \centering
  \includegraphics[width=1\textwidth]{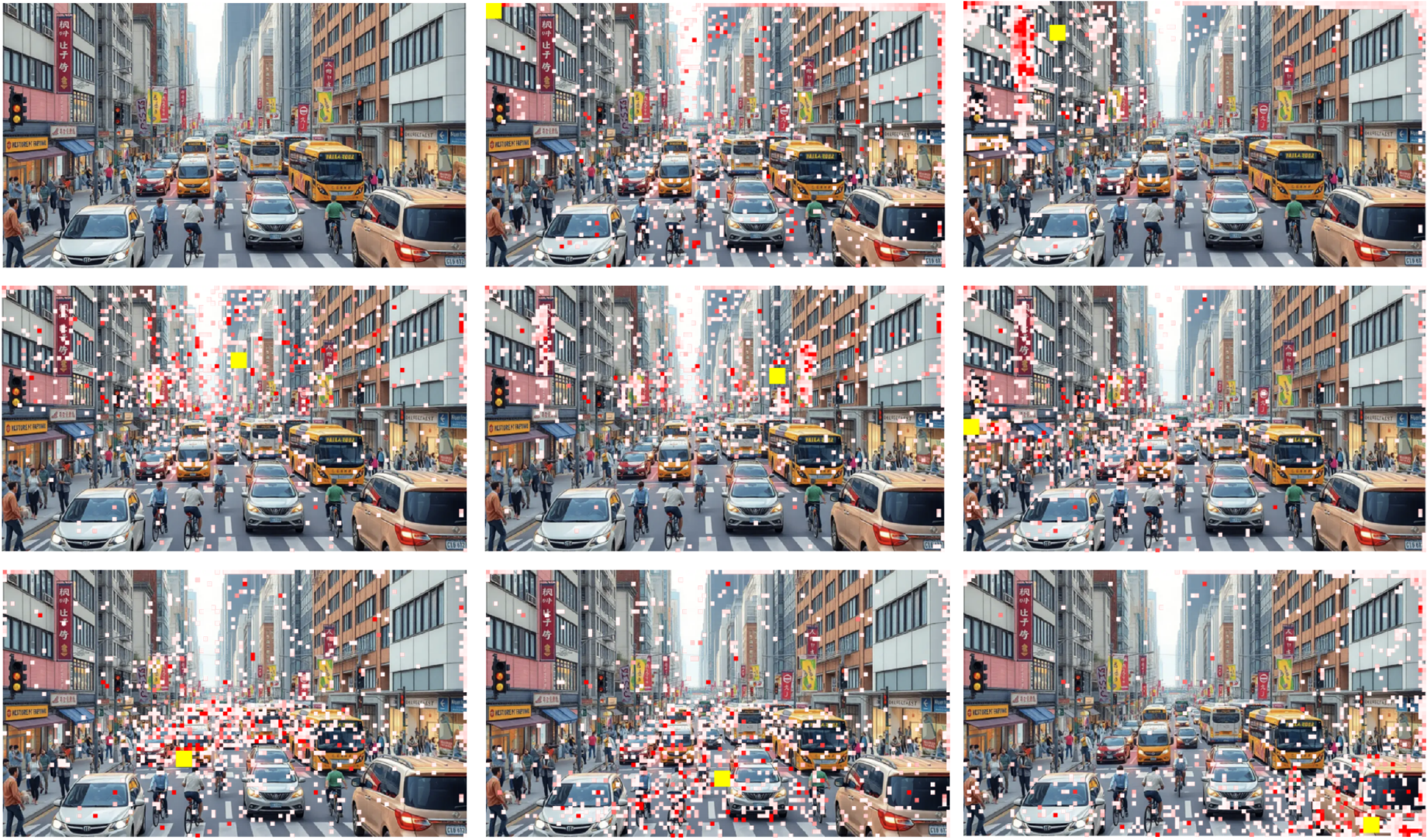}
  \label{fig:street}
\end{figure*}
\begin{figure*}[h]
  \centering
  \includegraphics[width=1\textwidth]{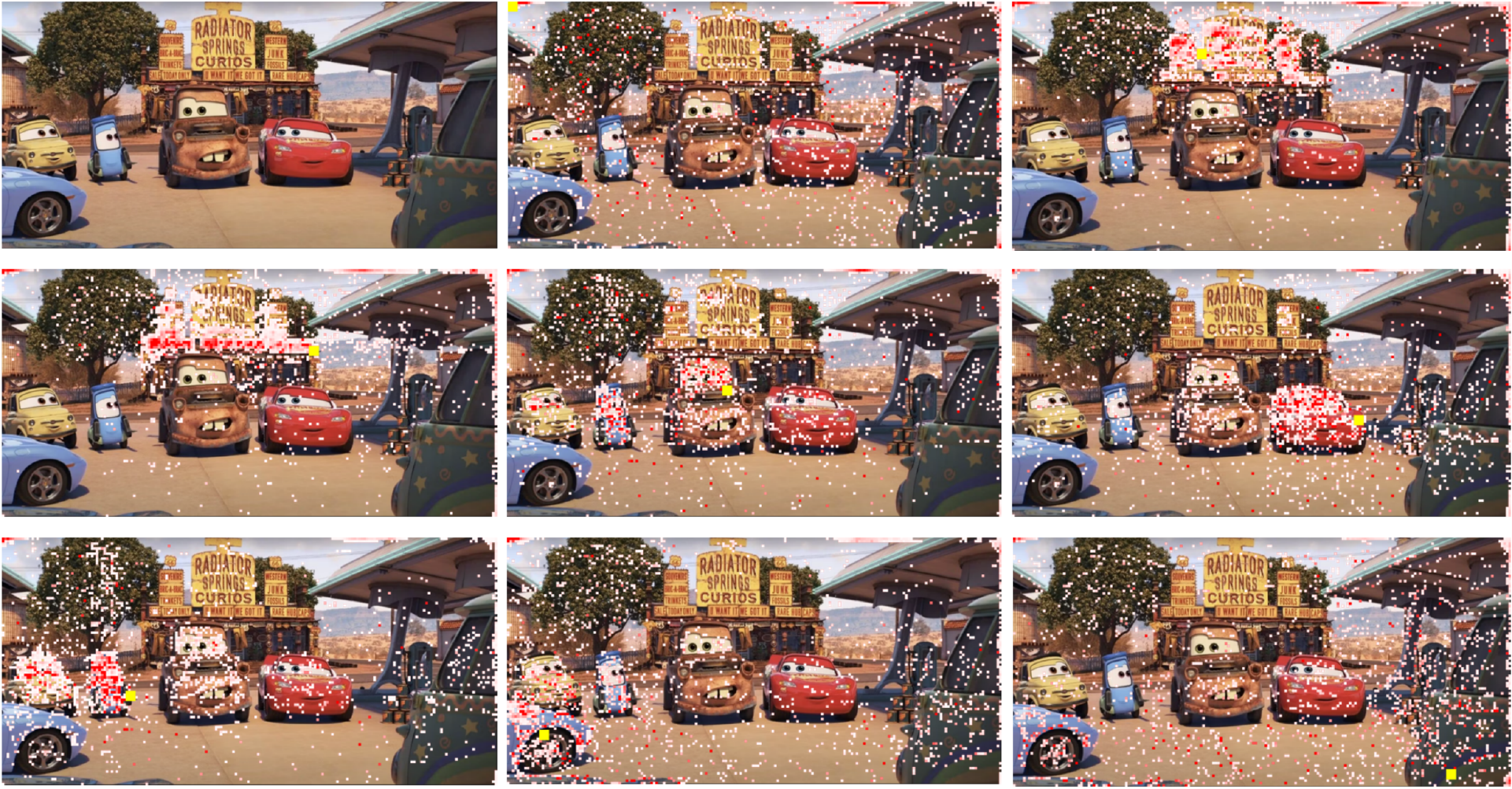}
  \caption{\textbf{Visualization of Position Encoding.} We initialize learnable standby tokens by uniformly sampling RoPE embeddings from the original sequence. We visualize their attention maps in the last ViT layer, marking assigned positions with a yellow square. For clarity, we display only the top 10\% of attention weights, where darker red indicates higher intensity. The results reveal a strong localization effect: standby tokens primarily absorb information from their neighboring patches.}
  \label{fig:cars}
\end{figure*}
\section{Visualization}

We visualize the position encoding mentioned in Sec.~3.3.1. For the newly introduced learnable standby tokens, we assign positional embeddings by uniformly sampling the RoPE encodings from the original sequence. In Fig.~\ref{fig:cars}, we visualize the attention distribution of these standby tokens towards other visual patches in the final layer of the ViT. We observe a distinct positional clustering effect (or spatial locality bias), where standby tokens tend to aggregate information from spatially adjacent tokens. This behavior aligns perfectly with our design expectations.

\begin{figure*}[h]
  \centering
  \includegraphics[width=0.99\textwidth]{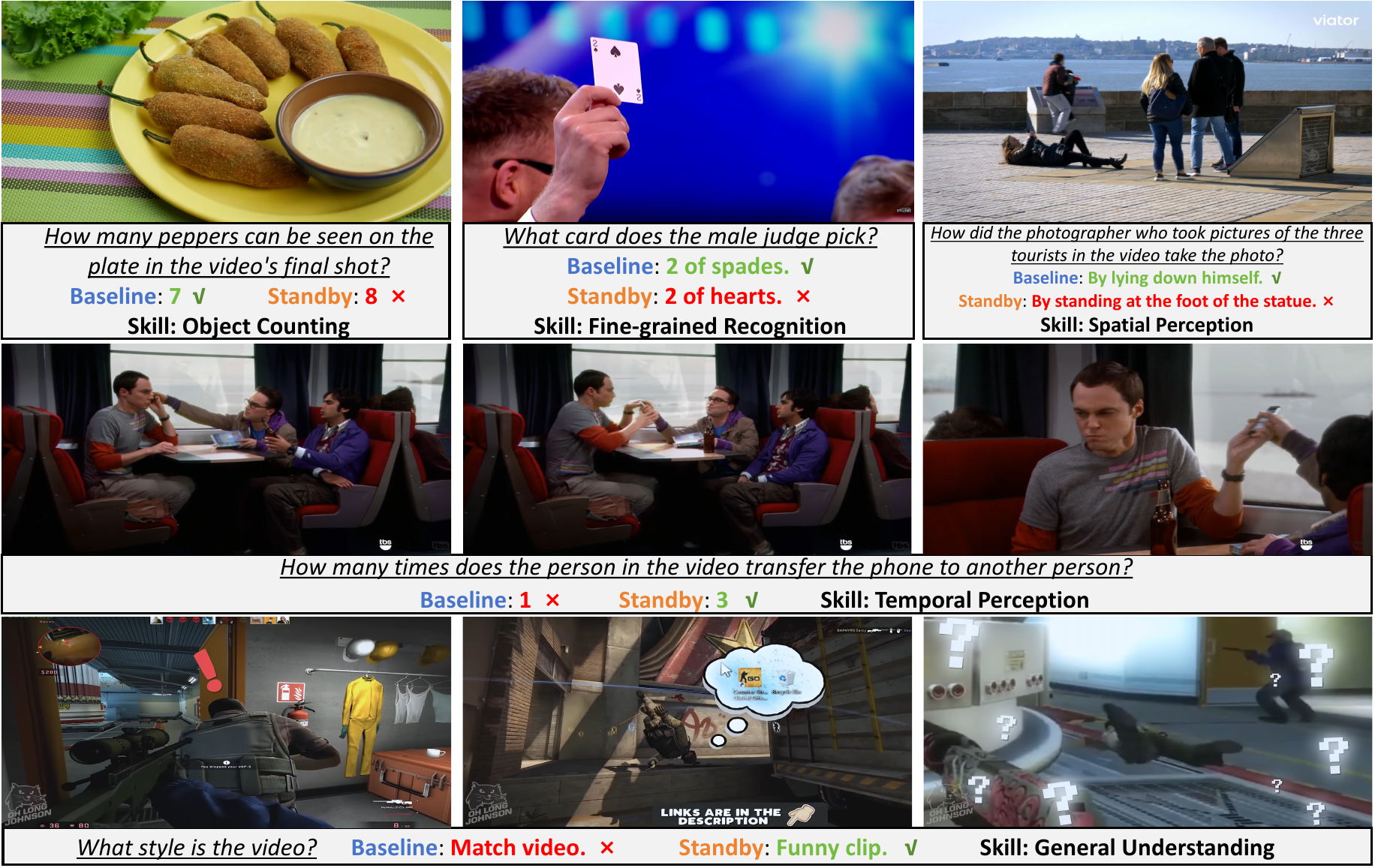}
  \caption{\textbf{Failure case analysis.} Standby mode fails on spatial or fine-grained perception while the baseline fails more on temporal and general understanding.}
  \label{fig:vis10}
\end{figure*}

\section{Failure Case Analysis}
We conduct a qualitative analysis on Video-MME in Fig.~\ref{fig:vis10}, comparing Baseline (64 frames) with Standby mode (128 frames). As shown in the figure below, many Standby failures ($>50$\%) are caused by deficits in spatial or fine-grained perception, whereas the Baseline fails more often on temporal and general understanding.

\section{Limitation \& Future Work}
In this work, we provided a comprehensive analysis of training dual-mode MLLMs and validated their effectiveness across both offline and streaming scenarios. However, the full potential of this dual-mode architecture remains under-explored. For instance, future training strategies could involve interleaved mode switching or utilizing the Standby mode to scale up the number of training frames. Ideally, the model should autonomously determine the appropriate inference mode via post-training strategies, potentially achieving frame-level precision.

Consider a long video understanding scenario: the model could first ingest densely sampled frames in Standby mode, then dynamically select keyframes to examine in Focus mode based on the specific query. This concept of 'thinking with videos' mirrors human cognitive patterns: skimming the video first and answering directly if the question is general, or revisiting specific segments based on memory if the query requires fine-grained details. We plan to prioritize exploring such complex reasoning patterns in future work. Notably, this form of adaptive visual thinking is unattainable without POINTS-Long’s dual-mode design, which uniquely enables reasoning over the entire video context. We hope this establishes a new direction for the field of visual reasoning.


\end{document}